\documentclass{article}

\usepackage{arxiv}
\usepackage{natbib}
\usepackage{bm}

\usepackage[utf8]{inputenc}
\usepackage[T1]{fontenc}
\usepackage{hyperref}
\usepackage{url}
\usepackage{booktabs}
\usepackage{amsfonts}
\usepackage{nicefrac}
\usepackage{microtype}
\usepackage{xcolor}

\usepackage{graphicx}
\usepackage{xspace}
\usepackage{mathtools}
\usepackage{amsmath}
\usepackage{amsthm}
\usepackage{amsfonts}
\usepackage{amssymb}
\usepackage{caption}
\usepackage{subcaption}
\usepackage{multirow}

\usepackage{algorithmic}
\usepackage{algorithm}
\usepackage{booktabs}
\usepackage{float}

\newtheorem{theorem}{Theorem}

\newtheorem{remark}{Remark}
\newtheorem{definition}{Definition}

\title{On Generalization and Distributional Update for Mimicking Observations with Adequate Exploration}

\author{
Yirui Zhou$^{1,2,}$\footnote{Equal contributions. }~~,Yunfei Jin$^{1,\ast}$, Xiaowei Liu$^{3,\ast}$, Xiaofeng Zhang$^4$, Yangchun Zhang$^{2,}$\thanks{Corresponding author. } \\
$^1$Shanghai Institute of Satellite Engineering, Shanghai, 201109, China \\
$^2$Department of Mathematics, College of Sciences, Shanghai University, Shanghai, 200444, China \\
$^3$School of Computer Science, University of Bristol, Bristol, BS8 1UB, UK \\
$^4$Newtouch Center for Mathematics of Shanghai University, Shanghai, 200444, China
}

\begin{document}
\maketitle

\begin{abstract}
Learning from observations (LfO) replicates expert behavior without needing access to the expert's actions, making it more practical than learning from demonstrations (LfD) in many real-world scenarios. However, directly applying the on-policy training scheme in LfO worsens the sample inefficiency problem, while employing the traditional off-policy training scheme in LfO magnifies the instability issue. This paper seeks to develop an efficient and stable solution for the LfO problem. Specifically, we begin by exploring the generalization capabilities of both the reward function and policy in LfO, which provides a theoretical foundation for computation. Building on this, we modify the policy optimization method in generative adversarial imitation from observation (GAIfO) with distributional soft actor-critic (DSAC), and propose the Mimicking Observations through Distributional Update Learning with adequate Exploration (MODULE) algorithm to solve the LfO problem. MODULE incorporates the advantages of (1) high sample efficiency and training robustness enhancement in soft actor-critic (SAC), and (2) training stability in distributional reinforcement learning (RL). Extensive experiments in MuJoCo environments showcase the superior performance of MODULE over current LfO methods. 
\end{abstract}

\noindent
{\it Keywords:} Learning from observations, Generalization properties, Distributional soft actor-critic

\section{Introduction}
Imitation learning (IL) \citep{pomerleau1991efficient,ng2000algorithms,syed2007a,ho2016generative}, a realm distinct from standard reinforcement learning (RL) \citep{puterman2014markov,sutton2018reinforcement}, does not rely on rewards provided by the environment. This characteristic makes IL well-suited for various real-world applications \citep{bhattacharyya2018multi,shi2019virtual,jabri2021robot}. The general IL paradigm leverages the guidance from expert demonstrations with information of both states and actions to mimic an outstanding policy \citep{abbeel2004apprenticeship,ho2016generative,kostrikov2020imitation}. According to the policy training strategy, IL is divided into two main schemes based on policy training strategy: on-policy and off-policy training. The on-policy scheme \citep{ho2016generative,chen2020on} is noted for its stability but requires a significant volume of samples. Conversely, the traditional off-policy scheme \citep{kostrikov2019discriminator,zhou2022generalization} is more sample-efficient but is at risk of higher training variance. 

In numerous practical contexts, such as camera systems, obtaining expert actions (like knowing the angle of rotation from the source to the target point cloud in camera scenarios) can be costly or even impractical. A more feasible and distinctive approach is gathering observational data without explicit action signals, known as learning from observations (LfO) \citep{torabi2018generative,yang2019imitation,torabi2019recent,zhu2020off,lee2021generalizable}. One feature of LfO is that it focuses on observing outcomes rather than replicating specific actions. This situation differs from the aforementioned learning from demonstrations (LfD) paradigm, which includes action information from experts. 

\begin{figure}[htbp]
\centering
\includegraphics[width=0.4\textwidth]{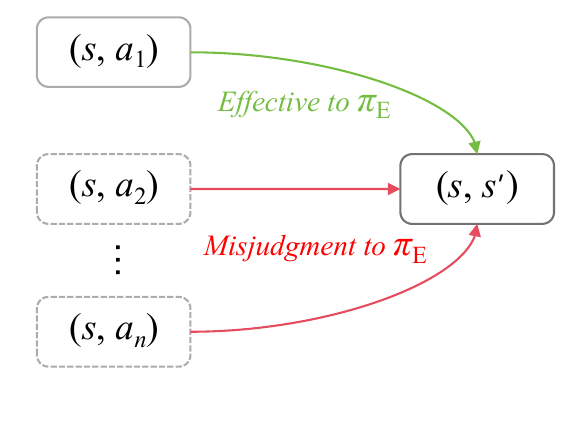}
\caption{An illustration of the misjudgment by the learned reward: Given $(s,s')$ as the expert's state $s$ and its next state $s'$ in LfO, which corresponds to the expert state-action pair $(s,a_{1})$ in LfD, misjudgment occurs when the agent might execute alternative actions $a_{i}$ ($i=2,\ldots,n$) that have the same transition dynamics leading to $s'$.}
\label{fig_off-policy_mistaken_misjudgment_example}
\end{figure}

We find that directly applying the two primary LfD policy training schemes in LfO tends to magnify their adverse effects. 
\begin{itemize}
\item Sample inefficiency problem in the on-policy scheme worsens with LfO. Since only state information is contained in the expert dataset, the learned reward is only able to distinguish whether the pair of the state and its next state $(s,s')$ is from the expert or the agent, which poses a forced wastage of the action information stored in the replay buffer \citep{lin1992self}, thereby decreasing the sample efficiency. 
\item Instability issue in the traditional off-policy scheme worsens with LfO. There exists a lack of training compatibility between the learned reward (depending on states $s,s'$) and the agent (depending on state $s$, action $a$) in LfO. For instance (Fig.\,\ref{fig_off-policy_mistaken_misjudgment_example}), the sequences $(s,a_{i},s'), i=1,\ldots,n$ generated by the agent will be uniformly recognized as approaching the expert. This increases the risk of misjudgment by the learned reward, leading to training instability. 
\end{itemize}
Therefore, only relying on on-policy or traditional off-policy training schemes hardly meets the requirements of satisfactory imitation performance (high sample efficiency with strong stability) for LfO. 

This paper starts by laying the theoretical foundation for computation in LfO. Specifically, we explore the generalization capabilities of both reward function and policy within the LfO problem \citep{torabi2018generative}:
\begin{align}
\hskip-0.3cm \mathop{\min}_{\pi \in \Pi}\mathop{\max}_{r\in \mathcal{R}}\mathbb{E}_{(s,s')\sim \mu_{\pi_{\rm E}}}\left[ r(s,s') \right] - \mathbb{E}_{(s,s')\sim \mu_{\pi}}\left[ r(s,s') \right], 
\label{lfo_problem}
\end{align}
where $\Pi$ and $\mathcal{R}$ represent the policy class and the reward function class, respectively. $\pi_{\rm E}$ denotes the expert policy and $\mu_{\pi}$ is the state transition distribution induced by the policy $\pi$ \citep{yang2019imitation,zhu2020off}. Consequently, we establish the generalization results of Eq. \eqref{lfo_problem}, which are based on the findings of generative adversarial imitation learning (GAIL) in \citep[Lemma 2]{xu2020error} and \citep[Theorem 1]{zhou2023distributional}. This enables us to handle the computation involved in LfO using GAIL's computational strategy. In particular, we modify the policy optimization method in generative adversarial imitation from observation (GAIfO) \citep{torabi2018generative} with distributional soft actor-critic (DSAC) \citep{ma2020dsac,duan2022distributional}, referred to as the Mimicking Observations through Distributional Update Learning with adequate Exploration (MODULE) algorithm. The benefit of MODULE is two-fold: 
\begin{itemize}
\item To enhance sample efficiency and operational simplicity (the insensitivity of the hyperparameter), MODULE incorporates the idea of a stochastic off-policy algorithm - soft actor-critic (SAC) \citep{haarnoja2018soft1,haarnoja2018soft2}, simultaneously avoiding the problem of drastic gradient changes in deterministic policies under GAN-based adversarial learning frameworks \citep{wang2024exploring}. Moreover, SAC employs a maximum entropy objective to ensure adequate exploration, which augments the robustness during the training process \citep{haarnoja2017reinforcement,haarnoja2018soft1}. 
\item To reduce the instability issue associated with the traditional off-policy training scheme, MODULE also leverages the technique of distributional RL. This approach is chosen for its superior ability to handle randomness (gain stability) and augment the detail in value function estimates through the modeling of return distributions \citep{bellemare2017distributional}. 
\end{itemize}
In short, the MODULE algorithm is capable of efficiently addressing the problem of Eq. \eqref{lfo_problem} in LfO. 

The remainder of this paper is arranged as outlined below. Section \ref{section_literature_review} provides a brief overview of the related work. Section \ref{section_preliminaries} describes the background on LfO and DSAC. The main generalization results for LfO are discussed in Section \ref{section_lfo_generalization}. Section \ref{section_MODULE} presents the MODULE algorithm as a solution to the LfO problem. Section \ref{section_evaluation} offers a detailed evaluation of MODULE's performance. Finally, Section \ref{section_conclusion} provides a summary and some discussions of this paper. 

\section{Literature review\label{section_literature_review}}
Our goal is to leverage methodologies for investigating LfD generalization properties to explore counterpart properties in LfO. Subsequently, we incorporate distributional RL into LfO to more accurately mimic expert observations. To this end, we review the existing literature on methodologies for analyzing LfD generalization properties (Section \ref{related_work_generalization_lfd}) and advancements in LfO research (Section \ref{related_work_lfo}).

\subsection{Generalization properties for learning from demonstrations\label{related_work_generalization_lfd}}
A body of research has delved into the generalization capability within the context of LfD \citep{chen2020on,xu2020error,zhou2023distributional,zhou2024generalization}. Specifically, \cite{chen2020on} showed that it is practicable to ensure the generalization of $\mathcal{R}$-distance by properly controlling the reward function in GAIL with parameterized rewards. In the follow-up study, under the metric of the neural network distance \citep{arora2017generalization}, the connection between generalization capabilities and expected return is established by analyzing the linear span of the discriminator class \citep{xu2020error}. \cite{zhou2023distributional} contributed insights into the policy class, leveraging the state-action distribution error to analyze the generalization of GAIL. Furthermore, \citep{zhou2024generalization} investigates the behavior of the generalization error bound within the reward transfer paradigm in transfer imitation learning (TIL), where state-action demonstration data is available exclusively in the source environment. Nevertheless, LfO generalization with respect to both the reward function class and the policy class within a single environmental setting remains an under-explored area in the literature.

\subsection{Learning from observations\label{related_work_lfo}}
The LfO framework \citep{torabi2018generative,yang2019imitation,torabi2019recent,zhu2020off,lee2021generalizable} tasks an agent to master tasks by learning solely from observational data, without access to the corresponding actions performed by the expert. This setup diverges from the general IL paradigm \citep{pomerleau1991efficient,ng2000algorithms,syed2007a,ho2016generative} by omitting the guidance of expert interventions, thereby escalating the difficulty for the agent to decipher and replicate the demonstrated behaviors. 

One effective strategy for tackling LfO is to estimate proxy rewards from the states encountered by both the learning agent and the expert, and then integrate these into a RL framework to guide the agent toward improving its performance by maximizing the total accumulated reward \citep{torabi2018generative,yang2019imitation,lee2021generalizable,jaegle2021imitation,liu2024imitation}. The majority of methods that can derive proxy rewards include optimal transport (OT) \citep{villani2009optimal} based approaches \citep{liu2024imitation,chang2024imitation}, techniques focused on estimating goal proximity \citep{lee2021generalizable,bruce2023learning}, and those employing the adversarial training mechanism \citep{torabi2018generative,yang2019imitation,sikchi2023ranking}.

\noindent \textbf{OT based approaches.} Recently, the concept of OT \citep{villani2009optimal} has been incorporated into the field of IL to label the proxy rewards \citep{liu2024imitation,chang2024imitation}. \cite{liu2024imitation} designed the automatic discount scheduling (ADS) mechanism to dynamically adjust the discount factor in RL, focusing on earlier rewards and progressively emphasizing later rewards. Observational off-policy Sinkhorn (OOPS) \citep{chang2024imitation} optimizes a reward function by reducing the discrepancy between the state trajectories of experts and learners, quantified through the Wasserstein distance. \\

\noindent \textbf{Goal proximity estimation.} \cite{lee2021generalizable} utilized the information of task progress readily and intuitively available in demonstration data, i.e., goal proximity, to derive a generalizable reward. In addition, \citep{bruce2023learning} introduces the explore like experts method, which employs expert data to approximate the monotonically increasing progress function through the temporal difference between frames that serves as an exploration reward, aiding the agent in efficiently navigating sparse-reward environments. \\

\noindent \textbf{Adversarial training.} A broad spectrum of studies adopt the mechanism of adversarial training to gain insights from expert observations \citep{torabi2018generative,sun2019provably,yang2019imitation,gangwani2020state,karnan2022adversarial,huang2024diffusion}. Expanding on the work of GAIL \citep{ho2016generative}, \cite{torabi2018generative} developed a model-free algorithm termed GAIfO and highlighted its role as a particular implementation within the general LfO framework. Forward adversarial imitation learning (FAIL) \citep{sun2019provably} transforms the LfO sequential learning issues into a series of independent two-player min-max games, which are resolved by employing no-regret online learning techniques. Besides, \cite{yang2019imitation} presented inverse-dynamics-disagreement-minimization (IDDM) to enhance LfO by better aligning the inverse dynamics models of the agent and the expert, effectively closing the divide towards LfD. Unlike direct imitation methods like GAIL and adversarial inverse reinforcement learning (AIRL) \citep{fu2018learning}, indirect imitation learning (I2L) \citep{gangwani2020state} utilizes an intermediary distribution to train a policy in a different MDP with distinct transition dynamics, based solely on expert states. In response to the challenge of high sample complexity partly stemming from the high-dimensional characteristic of video data, visual generative adversarial imitation from observation using a state observer (VGAIfO-SO) \citep{karnan2022adversarial} leverages a self-supervised state observer to distill high-dimensional visual information into lower-dimensional state representations. Further, \cite{huang2024diffusion} introduced diffusion imitation from observation (DIFO) that embeds a diffusion model within an adversarial imitation learning from observation setup to serve as a discriminator, customizing rewards for policy optimization. 

Among these methods, our MODULE algorithm hinges on adversarial training as its foundational methodology. This process develops a function capable of effectively distinguishing the data between the expert and the agent, thereby constructing the proxy reward. 

In a distinct area of LfO research, several approaches concentrate on model-based techniques, where the agent learns an inverse dynamics model (IDM) to infer the missing action information for the expert \citep{torabi2018behavioral,zhu2020off,candido2024mimicking}. Indeed, behavioral cloning from observation (BCO) \citep{torabi2018behavioral} can imitate policies without requiring access to demonstrator actions or post-demonstration interaction with the environment. However, BCO demonstrates poor performance with a learned IDM and fails to learn the consistent policy across both state-based and image-based experiments \citep{sikchi2024dual}. An alternative approach named off-policy learning from observations (OPOLO) \citep{zhu2020off} uses principled off-policy improvement and accelerates the learning process by employing an inverse action model to guide policy updates, thereby promoting distribution matching in the mode-covering perspective. Mimicking better by matching the approximate action distribution \citep{candido2024mimicking} incorporates an IDM within the framework of on-policy LfO. Our MODULE algorithm differs from model-based approaches, yet it has the potential to augment stability when incorporating the distributional update strategy with model-based components. 

\section{Preliminaries\label{section_preliminaries}}
The model of how the agent interacts with the environment is captured using a Markov decision process (MDP) characterized by the tuple $(\mathcal{S}, \mathcal{A}, \mathcal{P}, r_{gt}, \gamma)$ with the finite state space $\mathcal{S}$ and the action space $\mathcal{A}$. $\mathcal{P}: \mathcal{S}\times \mathcal{A}\times \mathcal{S}\rightarrow [0,1]$ is the transition dynamics, $r_{gt}: \mathcal{S}\times \mathcal{A}\rightarrow \mathbb{R}$ denotes the ground truth reward, and $\gamma$ represents the discount factor. A stochastic policy $\pi(a|s)$ assigns probabilities to actions $a\in \mathcal{A}$ based on the current state $s\in \mathcal{S}$. Following previous works \citep{xu2020error,yang2019imitation,zhu2020off}, we use the notations for the stationary distributions associated with $\pi$: $d_{\pi}(s)$ for the state, $\rho_{\pi}(s,a)$ for the state-action pair and $\mu_{\pi}(s,s')$ for the state transition. Their definitions are listed in Table \ref{table_stationary_distributions}. 

\begin{table*}[htbp]
\centering
\caption{Different stationary distributions. }
\label{table_stationary_distributions}
\begin{tabular}{llll} 
\toprule
Stationary distribution & Notation & Support & Definition \\
\midrule
State distribution & $d_{\pi}(s)$ & $\mathcal{S}$ & $(1-\gamma)\sum_{t=0}^{\infty}{\gamma ^{t}{\rm Pr}(s_{t}=s;\pi)}$ \\
State-action distribution & $\rho_{\pi}(s,a)$ & $\mathcal{S}\times \mathcal{A}$ & $(1-\gamma)\sum_{t=0}^{\infty}{\gamma ^{t}{\rm Pr}(s_{t}=s,a_{t}=a;\pi)}$ \\
State transition distribution & $\mu_{\pi}(s,s')$ & $\mathcal{S}\times \mathcal{S}$ & $\int_{\mathcal{A}}\rho_{\pi}(s,a)\mathcal{P}(s'|s,a)da$ \\
\bottomrule
\end{tabular}
\end{table*}

\subsection{Adversarial training in learning from observations\label{preliminaries_lfo}}
Due to the fact that only states rather than state-action pairs are provided as expert guidance, the challenge in LfO increases in comparison to LfD. The adversarial training problem in LfO is formulated as the min-max optimization problem in Eq. \eqref{lfo_problem}. It aims to find a policy within the policy class $\Pi$ that recovers the expert policy $\pi_{\rm E}$ using expert observation data. The learned reward is designed to distinguish between state transition pairs from expert observations and those produced by the agent. Simultaneously, the policy aims to generate interaction data that challenges the learned reward's ability to make this distinction, which corresponds to solving the following problem under a fixed reward $r\in \mathcal{R}$:
\begin{align*}
\mathop{\max}_{\pi \in \Pi}\mathbb{E}_{(s,s')\sim \mu_{\pi}}\left[ r(s,s') \right].
\end{align*}

\subsection{Distributional soft actor-critic\label{preliminaries_dsac}}
DSAC updates the policy by a maximum entropy RL objective under the reward function $r$ with the entropy temperature parameter $\alpha$: 
\begin{align*}
\pi^{\star}=\mathop{\arg\max}_{\pi}\mathbb{E}_{\pi}\left[ \sum_{t}{\gamma ^{t}\left( r(s_{t},a_{t})+\alpha \mathbb{H}(\pi(\cdot|s_{t})) \right)} \right]. 
\end{align*}
It is established on distributional soft policy iteration, which consists of distributional soft policy evaluation and distributional soft policy improvement \citep{ma2020dsac,duan2022distributional}. During the distributional soft policy evaluation stage, the distributional soft Bellman operator with respect to a fixed policy and the corresponding soft action-value distribution is denoted as follows: 
\begin{align*}
\mathcal{T}_{DS}^{\pi}Z(s,a): \overset{D}{=} r(s,a)+\gamma \big[Z(s',a')-\alpha \log \pi(a'|s') \big], s'\sim \mathcal{P}(\cdot|s,a),a'\sim \pi(\cdot|s'), 
\end{align*}
where $U:\overset{D}{=}V$ specifies identical probability distributions for random variables $U$ and $V$. Then, the distributional soft policy improvement stage updates the policy towards the exponential form of the soft action-value function: 
\begin{eqnarray}
\pi_{new}=\mathop{\arg \min}_{\pi' \in \Pi}D_{KL}\left(\pi'\left(\cdot|s_{t}\right) \bigg\| \frac{\exp \left(\frac{1}{\alpha} Q_{soft}^{\pi_{old}}\left(s_{t},\cdot\right)\right)}{\Delta^{\pi_{old}}(s_{t})}\right), 
\label{SAC_policy_KL}
\end{eqnarray}
where $\Delta^{\pi_{old}}$ is the normalization factor for the distribution. By iteratively applying distributional soft policy evaluation and distributional soft policy improvement, DSAC converges toward optimality. 

\section{Generalization properties of LfO\label{section_lfo_generalization}}
In this section, we analyze the generalization capability for the reward function and the policy learned by the problem of Eq. \eqref{lfo_problem} in LfO, respectively. 

\subsection{Generalization for the reward function}
Analogous to the $\mathcal{R}$-distance over $s,a$ in \citep[Definition 2]{chen2020on}, we first define the LfO version - the LfO reward distance and its empirical version. 

\begin{definition}[LfO reward distance]\label{definition_LfO_reward_distance}
For a class of reward functions $\mathcal{R}$ with respect to $s,s'$, there exists $r\in \mathcal{R}$ such that $r(s,s')\equiv 0$. The LfO reward distance between two state transition distributions $\mu_{\pi}$ and $\mu_{\pi'}$ is defined as
\begin{align*}
d_{\mathcal{R}}^{\rm LfO}(\mu_{\pi}, \mu_{\pi'}) 
=\sup_{r\in \mathcal{R}}\left\{ \mathbb{E}_{(s,s')\sim \mu_{\pi}}{[r(s,s')]}-\mathbb{E}_{(s,s')\sim \mu_{\pi'}}{[r(s,s')]}\right\}. 
\end{align*}
\end{definition}

To make sure that the LfO reward distance remains non-negative, the zero function is assumed to be incorporated into the function class $\mathcal{R}$, as described in Definition \ref{definition_LfO_reward_distance}. This concept of distance is referred to as an integral probability metric (IPM) \citep{muller1997integral} over state transition distributions. 

We denote $d_{\mathcal{R}}^{\rm LfO}(\hat{\mu}_{\pi}, \hat{\mu}_{\pi'})$ as the empirical LfO reward distance, where $\hat{\mu}_{\pi}$ and $\hat{\mu}_{\pi'}$ are the empirical counterparts of $\mu_{\pi}$ and $\mu_{\pi'}$, respectively, estimated from $n$ samples. In practice, Eq. \eqref{lfo_problem} achieves the minimization of the empirical LfO reward distance between $\hat{\mu}_{\pi_{\rm E}}$ and $\hat{\mu}_{\pi}$. We now introduce a theorem that elucidates the generalization capability for the reward function of LfO, which is based on the result for GAIL in \citep[Lemma 2]{xu2020error}. 

\begin{theorem}[LfO generalization for the reward function]\label{theorem_lfo_generalization_reward}
For a uniformly bounded reward function class $\mathcal{R}$ with respect to $s,s'$, i.e., for any $r\in \mathcal{R}$, $\max_{s,s'}|r(s,s')|\leq B_{r}$, and the policy $\pi_{\rm I}$ learned by Eq. \eqref{lfo_problem} satisfies
\begin{align*}
d_{\mathcal{R}}^{\rm LfO}(\hat{\mu}_{\pi_{\rm E}}, \hat{\mu}_{\pi_{\rm I}})\leq \inf_{\pi \in \Pi}d_{\mathcal{R}}^{\rm LfO}(\hat{\mu}_{\pi_{\rm E}}, \hat{\mu}_{\pi})+\hat{\epsilon}_{r},
\end{align*}
then for all $\delta \in (0,1)$, with probability at least $1-\delta$, we have that
\begin{align*}
d_{\mathcal{R}}^{\rm LfO}(\mu_{\pi_{\rm E}},\mu_{\pi_{\rm I}})\leq \inf_{\pi \in \Pi}d_{\mathcal{R}}^{\rm LfO}(\hat{\mu}_{\pi_{\rm E}}, \hat{\mu}_{\pi})+2\hat{\mathfrak{R}}_{\mu_{\pi_{\rm E}}}^{(n)}(\mathcal{R}) +2\hat{\mathfrak{R}}_{\mu_{\pi_{\rm I}}}^{(n)}(\mathcal{R})+12B_{r}\sqrt{\frac{\log(4/\delta)}{2n}}+\hat{\epsilon}_{r}. 
\end{align*}
\end{theorem}

For detailed proof, please refer to Appendix \ref{proof_theorem_lfo_generalization_reward}. Theorem \ref{theorem_lfo_generalization_reward} suggests that a well-controlled reward function class $\mathcal{R}$ ensures the generalization of the LfO reward distance, and subsequently guarantees the generalization of problem Eq. \eqref{lfo_problem} throughout the training phase. It is worth noting that \citep[Theorem 2]{xu2020error} extends the generalization results under the ground truth reward $r_{gt}$ (the evaluation phase) by establishing its connection with the linear span of the trained function class. However, it is not reasonable to connect the state-action ground truth reward with the reward function class $\mathcal{R}$ regarding the state transition in LfO. We will explore this issue in the future. 

\subsection{Generalization for the policy}
Followed by the definition of state-action distribution error in \citep[Definition 1]{zhou2023distributional}, we characterize the state transition distribution error as follows. 

\begin{definition}[State transition distribution error]\label{definition_state_transition_distribution_error}
For a class of policies $\Pi$, an expert policy $\pi_{\rm E}$ and a fixed non-negative reward $r$, the state transition distribution error is defined as
\begin{align*}
\bm{e}(C_{r}\mu_{\pi_{\rm E}},C_{r}\mu_{\Pi})
=C_{r}\inf_{\pi \in \Pi}\left\{ \mathbb{E}_{(s,s')\sim \mathbb{R}}\left[ \mu_{\pi_{\rm E}}(s,s')-\mu_{\pi}(s,s') \right] \right\}, 
\end{align*}
where $\mu_{\Pi}$ represents the functional class of state transition distributions induced by $\Pi$, $C_{r}=\sum_{(s,s')\in \mathcal{S}\times \mathcal{S}}r(s,s')$ is the state transition reward coefficient and $\mathbb{R}(s,s')=r(s,s')/C_{r}$ is the state transition distribution normalized by the reward $r(s,s')$. 
\end{definition}

\begin{remark}
In practice, the shape of the learned reward $r$ is diverse, with possible value ranges including $(-\infty,+\infty)$, $[0,+\infty)$, $(-\infty,0]$, $[0,1]$, etc \citep{wang2021reward}. Here, we use a non-negative $r$ as a representative example to investigate the generalization property for the policy in LfO. 
\end{remark}

Note that
\begin{align*}
C_{r}\mathbb{E}_{(s,s')\sim \mathbb{R}}\left[ \mu_{\pi}(s,s') \right]=\mathbb{E}_{(s,s')\sim \mu_{\pi}}\left[ r(s,s') \right]. 
\end{align*}
Hence, Definition \ref{definition_state_transition_distribution_error} fundamentally quantifies the minimum error in expected return between the expert policy and the policy class: 
\begin{align*}
\bm{e}(C_{r}\mu_{\pi_{\rm E}},C_{r}\mu_{\Pi}) =\mathbb{E}_{(s,s')\sim \mu_{\pi_{\rm E}}}\left[ r(s,s') \right] - \sup_{\pi \in \Pi}\left\{ \mathbb{E}_{(s,s')\sim \mu_{\pi}}\left[ r(s,s') \right] \right\},
\end{align*}

In practice, Eq. \eqref{lfo_problem} maximizes the empirical state transition distribution error
\begin{align*}
\bm{e}(\hat{c}_{r}^{(m)}\mu_{\pi_{\rm E}},\hat{c}_{r}^{(m)}\mu_{\Pi})
=\hat{c}_{r}^{(m)}\inf_{\pi \in \Pi}\left\{ \mathbb{E}_{(s,s')\sim \hat{\mathbb{R}}}\left[ \mu_{\pi_{\rm E}}(s,s')-\mu_{\pi}(s,s') \right] \right\}, 
\end{align*}
where $\hat{c}_{r}^{(m)}=\sum_{i=1}^{m}r(s^{(i)},s'^{(i)})$ is the empirical state transition reward coefficient and $\hat{\mathbb{R}}(s^{(i)},s'^{(i)})=r(s^{(i)},s'^{(i)})/\hat{c}_{r}^{(m)}$ denotes the empirical counterpart of $\mathbb{R}$ with $m$ samples. Next, we provide the generalization capability for the policy of LfO, which is based on the consequence of GAIL in \citep[Theorem 1]{zhou2023distributional}. 

\begin{theorem}[LfO generalization for the policy]\label{theorem_lfo_generalization_policy}
Given the reward $r_{\rm I}$ learned by Eq. \eqref{lfo_problem} and a policy class $\Pi$, which satisfies for all $\pi \in \Pi$, 
\begin{align*}
\max_{s,s'}\left\{ C_{r_{\rm I}}\mu_{\pi}(s,s'),C_{r_{\rm I}}\mu_{\pi_{\rm E}}(s,s') \right\}\leq B_{\Pi}. 
\end{align*}
Assume
$$
\bm{e}(\hat{c}_{r_{\rm I}}^{(m)}\mu_{\pi_{\rm E}},\hat{c}_{r_{\rm I}}^{(m)}\mu_{\Pi}) \geq \sup_{r\in \mathcal{R}}\left\{ \bm{e}(\hat{c}_{r}^{(m)}\mu_{\pi_{\rm E}},\hat{c}_{r}^{(m)}\mu_{\Pi})\right\}-\hat{\epsilon}_{\pi}\geq 0,
$$
then for all $\delta \in (0,1)$, with probability at least $1-\delta$, we have that
\begin{align*}
\bm{e}(C_{r_{\rm I}}\mu_{\pi_{\rm E}},C_{r_{\rm I}}\mu_{\Pi})\geq \sup_{r\in \mathcal{R}}\left\{ \bm{e}(\hat{c}_{r}^{(m)}\mu_{\pi_{\rm E}},\hat{c}_{r}^{(m)}\mu_{\Pi}) \right\} -2\hat{\mathfrak{R}}_{\mathbb{R}_{\rm I}}^{(m)}(C_{r_{\rm I}}\mu_{\Pi})-8B_{\Pi} \sqrt{\frac{\log(3/\delta)}{2m}} -\hat{\epsilon}_{\pi}. 
\end{align*}
\end{theorem}

For detailed proof, please refer to Appendix \ref{proof_theorem_lfo_generalization_policy}. Theorem \ref{theorem_lfo_generalization_policy} implies that with effective control over the class of policies $\Pi$, there is an assurance of generalization in the state transition distribution error, consequently ensuring the generalization of problem Eq. \eqref{lfo_problem} during the training process. 

Building upon Theorem \ref{theorem_lfo_generalization_reward} and Theorem \ref{theorem_lfo_generalization_policy}, the generalization of LfO with adversarial learning (as presented in problem Eq. \eqref{lfo_problem}) is possessed. Subsequently, in the following section, we will introduce a specific computational methodology for LfO - the MODULE algorithm. 

\section{MODULE algorithm\label{section_MODULE}}
By the generalization of the reward function and the policy, we are able to explore the LfO computational properties (Eq. \eqref{lfo_problem}) via approximating them with the parameterizations $r_{\phi}$ and $\pi_{\theta}$. Specifically, $\mathcal{R}$ and $\Pi$ can be assigned by specific function classes \citep{chen2020on,zhou2023distributional}, e.g., reproducing kernel Hilbert space \citep{ormoneit2002kernel} or neural networks \citep{lecun2015deep}. Then the LfO problem Eq. \eqref{lfo_problem} evolves into
\begin{align}
\hskip-0.7cm \mathop{\min}_{\theta}\mathop{\max}_{\phi}\mathbb{E}_{(s,s')\sim \mu_{\pi_{\rm E}}}\left[ r_{\phi}(s,s') \right] - \mathbb{E}_{(s,s')\sim \mu_{\pi_{\theta}}}\left[ r_{\phi}(s,s') \right], 
\label{paramterization_lfo_problem}
\end{align}

Owing to the fact that the problem in Eq. \eqref{paramterization_lfo_problem} does not possess a convex-concave structure, motivated by \citep{chen2020on,zhou2023distributional}, the min-max optimization problem is slightly modified by
\begin{align}
\mathop{\min}_{\theta}\mathop{\max}_{\phi}\mathbb{E}_{(s,s')\sim \mu_{\pi_{\rm E}}}\left[ r_{\phi}(s,s') \right] - \mathbb{E}_{(s,s')\sim \mu_{\pi_{\theta}}}\left[ r_{\phi}(s,s') \right] -\alpha \mathbb{H}(\pi_{\theta})-\frac{\mu}{2}\|\phi\|_{2}^{2},
\label{modified_lfo_problem}
\end{align}
where $\mathbb{H}(\pi_{\theta})$ is the entropy of the policy $\pi_{\theta}$, and $\alpha,\mu$ are tuning parameters. This modification maximizes the entropy of the policy as much as possible, which encourages adequate exploration of the agent and enhances the policy training. 

To solve the problem in Eq. \eqref{modified_lfo_problem}, our MODULE algorithm alternates between updating the policy parameter $\theta$ and the reward function parameter $\phi$. For the update of the reward function parameter $\phi$, the gradient descent method is directly applied with the loss function
\begin{align}
L_{r}(\phi)=
&-\mathbb{E}_{(s,s')\sim \mathcal{D}^{\rm E}}\left[ r_{\phi}(s,s') \right] + \mathbb{E}_{(s,s')\sim \mathcal{D}^{\rm I}}\left[ r_{\phi}(s,s') \right] +\frac{\mu}{2}\|\phi\|_{2}^{2}, 
\label{objective_reward_phi}
\end{align}
where $\mathcal{D}^{\rm E}$ is the set of expert observations and $\mathcal{D}^{\rm I}$ denotes the replay buffer \citep{lin1992self}. For the update of the policy parameter $\theta$, its optimization problem aims to maximize both expected return under $r_{\phi}$ and the policy entropy, i.e., 
\begin{align*}
\mathop{\max}_{\theta} \mathbb{E}_{(s,s')\sim \mu_{\pi_{\theta}}}\left[ r_{\phi}(s,s') \right] + \alpha \mathbb{H}(\pi_{\theta}),
\end{align*}
which is essentially the objective of maximum entropy RL. Driven by this, our MODULE algorithm is designed to combine SAC (which adjusts entropy automatically) with the technique of distributional RL during policy updates. Particularly, (1) SAC, grounded in the principle of maximum entropy RL, guarantees that the learned policy accounts for all beneficial actions \citep{haarnoja2018soft2}. (2) Distributional RL, known for its powerful ability to model the distribution over returns and capture randomness, improves computational stability \citep{bellemare2017distributional}. 

Regarding (1), the value function network parameter $w$, the policy network parameter $\theta$ and the entropy temperature parameter $\alpha$ are updated in turns. Specifically, 
\begin{itemize}
\item \textbf{The objective of the value function network parameter $w$} is to minimize the soft Bellman residual \citep{haarnoja2018soft1,haarnoja2018soft2}, i.e., minimizing
\begin{eqnarray*}
J_{Q}(w)=\mathbb{E}_{(s_{t},a_{t})\sim \mathcal{D}^{\rm I}}\Big[ \frac{1}{2}\big( Q_{w}^{\rm soft}(s_{t},a_{t})-(r_{\phi}(s_{t},s_{t+1}) +\gamma \mathbb{E}_{s_{t+1}\sim \mathcal{P}(\cdot|s_{t},a_{t})}[V_{\bar{w}}^{\rm soft}(s_{t+1})]) \big)^{2} \Big], 
\end{eqnarray*}
where $\bar{w}$ is employed to compute the target value for training stability \citep{mnih2015human}, and 
$$
\hskip0.1cm V_{\bar{w}}^{\rm soft}(s_{t})=\mathbb{E}_{a_{t}\sim \pi_{\theta}}[Q_{\bar{w}}^{\rm soft}(s_{t},a_{t})-\alpha \log(\pi_{\theta}(a_{t}|s_{t}))]. 
$$
In practice, two value function networks with parameters $w_{1},w_{2}$ are used to relieve the overestimation problem \citep{fujimoto2018addressing}. 

\item \textbf{The objective of the policy network parameter $\theta$} is to minimize the expected Kullback-Leibler (KL) divergence in Eq. \eqref{SAC_policy_KL} \citep{haarnoja2018soft1,haarnoja2018soft2}, i.e., minimizing
\begin{align}
\hskip-0.9cm J_{\pi}(\theta)=\mathbb{E}_{s_{t}\sim \mathcal{D}^{\rm I}}\big[ \mathbb{E}_{a_{t}\sim \pi_{\theta}}[ \alpha \log \pi_{\theta}(a_{t}|s_{t})-Q_{w}^{\rm soft}(s_{t},a_{t})] \big]. 
\label{objective_policy_theta}
\end{align}

\item \textbf{The objective of the entropy temperature parameter $\alpha$} is to minimize the following objective: 
\begin{align}
\hskip-0.3cm J(\alpha)=\mathbb{E}_{a_{t}\sim \pi_{\theta}}\big[ -\alpha \log\pi_{\theta}(a_{t}|s_{t})-\alpha H_{0}\big], 
\label{objective_entropy_alpha}
\end{align}
where $H_{0}$ represents the desired minimum expected entropy, often set as the negative dimensionality of the action space $\mathcal{A}$ \citep{haarnoja2018soft2}. 
\end{itemize}

Regarding (2), we integrate the distributional RL technique into SAC, i.e., DSAC \citep{ma2020dsac,duan2022distributional}, as the policy optimization method in our MODULE algorithm. Two quantile value networks operate in place of the original two value function networks while still utilizing $w_{1}$ and $w_{2}$ as parameters. We detail the update of the quantile value network parameters and the computation of the risk soft action-value (used in Eq. \eqref{objective_policy_theta}) as follows. Specifically, 
\begin{itemize}
\item \textbf{The objective of the quantile value network parameters $w_{1},w_{2}$} is to minimize the weighted pairwise Huber regression loss across various quantile fractions \citep{ma2020dsac}
\begin{align}
J_{Z}(w)=\sum_{i=0}^{M-1}\sum_{j=0}^{M-1}\left(\tau_{i+1}-\tau_{i}\right) \rho_{\hat{\tau}_{j}}^{\kappa}\left(\delta_{ij}^{t}\right), 
\label{objective_quantile_w}
\end{align}
where $M$ is the size of the quantile fraction ensemble, $\{\tau_{i}\}_{i=1}^{M}$ and $\{\tau_{j}\}_{i=1}^{M}$ are the quantile fraction ensembles, which can be generated by quantile regression DQN (QR-DQN) \citep{dabney2018distributional}, implicit quantile networks (IQN) \citep{dabney2018implicit} or fully parameterized quantile function (FQF) \citep{yang2019fully}. $\hat{\tau}_{i}=(\tau_{i}+\tau_{i+1})/2$, $\hat{\tau}_{j}=(\tau_{j}+\tau_{j+1})/2$, 
\begin{align*}
\delta_{ij}^{t}=r_{t+1}+\gamma\big[ Z_{\hat{\tau}_{i},\bar{w}}\left(s_{t+1}, a_{t+1} \right) -\alpha \log \pi_{\bar{\theta}}\left(a_{t+1}| s_{t+1}\right) \big]-Z_{\hat{\tau}_{j},w}\left(s_{t}, a_{t} \right), 
\end{align*}
$\bar{\theta}$ is the target policy network parameter, and
\begin{align*}
\rho_{\tau}^{\kappa}\left(\delta_{ij}\right)=\left|\tau-\mathbb{I}\left\{ \delta_{i j}<0\right\}\right| \frac{\mathcal{L}_{\kappa}\left(\delta_{ij}\right)}{\kappa}, \end{align*}

\begin{align*}
\hskip-0.1cm \mathcal{L}_{\kappa}\left(\delta_{i j}\right)
=
\begin{cases}
~~~~~~~~~\frac{1}{2} \delta_{ij}^{2}, & \text{if}\left|\delta_{i j}\right| \leq \kappa, \\ 
\kappa\left(\left|\delta_{ij}\right|-\frac{1}{2} \kappa\right), & \text{otherwise}, 
\end{cases}
\end{align*}
where $\kappa$ is the threshold of the Huber quantile regression loss \citep{huber1964robust}. 

\item \textbf{The computation of the risk soft action-value \citep{ma2020dsac}} is performed by
\begin{align}
Q_{w}^{\rm soft}(s,a)&=\min_{i=1,2} Q_{w_{i}}^{\rm soft}(s,a), \notag\\
Q_{w_{i}}^{\rm soft}(s,a)&=\Psi \left[ Z_{\tau,w_{i}}(s,a) \right],
\label{soft_Q_computation}
\end{align}
where $\Psi:\mathcal{Z}\rightarrow \mathbb{R}$ is the risk measure function. Except for risk-neutral, common risk measures with parameter $\beta$, such as mean-variance \citep{sobel1982variance,tamar2012policy,prashanth2016variance}, value-at-risk (VaR) \citep{prashanth2018risk} and distorted expectation \citep{ma2020dsac}, are listed in Table \ref{table_risk_measure}. Particularly, popular distorted expectations, including cumulative probability weighting parameterization (CPW) \citep{tversky1992advances}, Wang's approach \citep{wang2000class} and conditional value at risk (CVaR) \citep{chow2015risk}, are detailed in Table \ref{table_distorted_expectations}, where $\Phi$ is the standard Normal cumulative distribution function (CDF). 
\end{itemize}

\begin{table}[htbp]
\centering
\caption{Function shapes of risk measures. }
\label{table_risk_measure}
\begin{tabular}{ll} 
\toprule
Risk measure & Function shape \\
\midrule
Risk-neutral & $\Psi(\cdot)=\mathbb{E}[\cdot]$ \\
Mean-variance & $\Psi(Z)=\mathbb{E}[Z]-\beta \sqrt{\mathbb{V}[Z]}$ \\
VaR & ${\rm VaR}_{\beta}(Z)=\min_{z}\left\{ z|F_{Z}(z)>\beta \right\}$ \\
Distorted expectation & $\Psi(Z)=\int_{0}^{1}F_{Z}^{-1}(\tau)dg(\tau)$ \\
\bottomrule
\end{tabular}
\end{table}

\begin{table}[htbp]
\centering
\caption{Function shapes of common distorted expectations. }
\label{table_distorted_expectations}
\begin{tabular}{ll} 
\toprule
Distorted expectation & Function shape of $g(\tau)$ \\
\midrule
CPW & $\tau^{\beta}/\left( \tau^{\beta}+(1-\tau)^{\beta} \right)^{1/\beta}$ \\
Wang & $\Phi(\Phi^{-1}(\tau)+\beta)$ \\
CVaR & $\min \left\{ \tau/\beta,1\right\}$ \\
\bottomrule
\end{tabular}
\end{table}

Overall, the procedure of our MODULE algorithm is organized in Algorithm \ref{alg_MODULE}. 

\begin{algorithm*}[htbp]
\caption{Mimicking Observations through Distributional Update Learning with adequate Exploration (MODULE)}
\begin{algorithmic}[1]
\STATE \textbf{Input:} Expert observations $\mathcal{D}^{\rm E}$, $\phi$, $\alpha$, $w_{1},w_{2}$, $\theta$, learning rates $\eta_{\phi}, \eta_{w}, \eta_{\theta}, \eta_{\alpha}$ and $\iota$
\STATE Initialize the target networks: $\bar{w}_{1}=w_{1}, \bar{w}_{2}=w_{2}$, $\bar{\theta}=\theta$
\STATE Initialize an empty replay buffer $\mathcal{D}^{\rm I}\leftarrow \varnothing$
\STATE \textbf{for} each iteration \textbf{do}
\STATE \hspace{0.5cm} Collect samples with the policy $\pi_{\theta}$, and then store the transitions in the replay buffer $\mathcal{D}^{\rm I}$
\STATE \hspace{0.5cm} \textbf{for} each reward updating step \textbf{do}
\STATE \hspace{1cm} Update $\phi$ by Eq. \eqref{objective_reward_phi} with samples from $\mathcal{D}^{\rm E}$ and $\mathcal{D}^{\rm I}$: $\phi \leftarrow \phi - \eta_{\phi}\nabla_{\phi}L_{r}(\phi)$
\STATE \hspace{0.5cm} \textbf{end for}
\STATE \hspace{0.5cm} \textbf{for} each policy updating step \textbf{do}
\STATE \hspace{1cm} Sample the transitions from $\mathcal{D}^{\rm I}$
\STATE \hspace{1cm} Update $w_{1},w_{2}$ by Eq. \eqref{objective_quantile_w} with the sampled transitions: $w_{i}\leftarrow w_{i}-\eta_{w}\nabla_{w}J_{Z}(w_{i})$ for $i\in \{1,2\}$
\STATE \hspace{1cm} Compute the risk soft action-value by Eq. \eqref{soft_Q_computation}
\STATE \hspace{1cm} Update $\theta$ by Eq. \eqref{objective_policy_theta} with the sampled transitions: $\theta \leftarrow \theta - \eta_{\theta}J_{\pi}(\theta)$
\STATE \hspace{1cm} Update $\alpha$ by Eq. \eqref{objective_entropy_alpha} with the sampled transitions: $\alpha \leftarrow \alpha - \eta_{\alpha}\nabla_{\alpha}J(\alpha)$
\STATE \hspace{1cm} Update target network weights: $\bar{w}_{i}\leftarrow \iota w_{i}+(1-\iota)\bar{w}_{i}$ for $i\in \{1,2\}$, $\bar{\theta}\leftarrow \iota \theta +(1-\iota)\bar{\theta}$
\STATE \hspace{0.5cm} \textbf{end for}
\STATE \textbf{end for}
\end{algorithmic}
\label{alg_MODULE}
\end{algorithm*}

\section{Evaluation\label{section_evaluation}}
In this section, we perform experiments in three MuJoCo environments \citep{todorov2012mujoco}: Hopper-v2, Walker2d-v2 and HalfCheetah-v2. These experiments aim to validate MODULE's performance, with a detailed experimental setup provided in Section \ref{section_experiment_setup}. Furthermore, Section \ref{section_experiment_others} compares MODULE with various advanced LfO methods, while Section \ref{section_experiment_ablation} examines the impact of different quantile fraction generation methods and risk measure functions on MODULE. 

\subsection{Experimental setup\label{section_experiment_setup}}
Following the conventions of commonly used IL experiments \citep{zhou2022generalization,zhou2023distributional,wang2024exploring}, we generate the expert dataset for each environment using an expert policy trained with a specific model-free algorithm - SAC \citep{haarnoja2018soft2}. The observation data, obtained from this expert policy with a 0.01 standard deviation, comprises $10^{6}$ state-action pairs. The average returns of the expert observation data are 3433, 3509 and 9890 for Hopper-v2, Walker2d-v2 and HalfCheetah-v2, respectively. In the forthcoming experiments, we replicate our trials 10 times with a uniform training configuration, varying only the random seeds. 

\subsection{Comparative experiments\label{section_experiment_others}}
\begin{figure*}[htbp]
\centering
\begin{minipage}{0.32\linewidth}
\vspace{1pt}
\includegraphics[width=\textwidth]{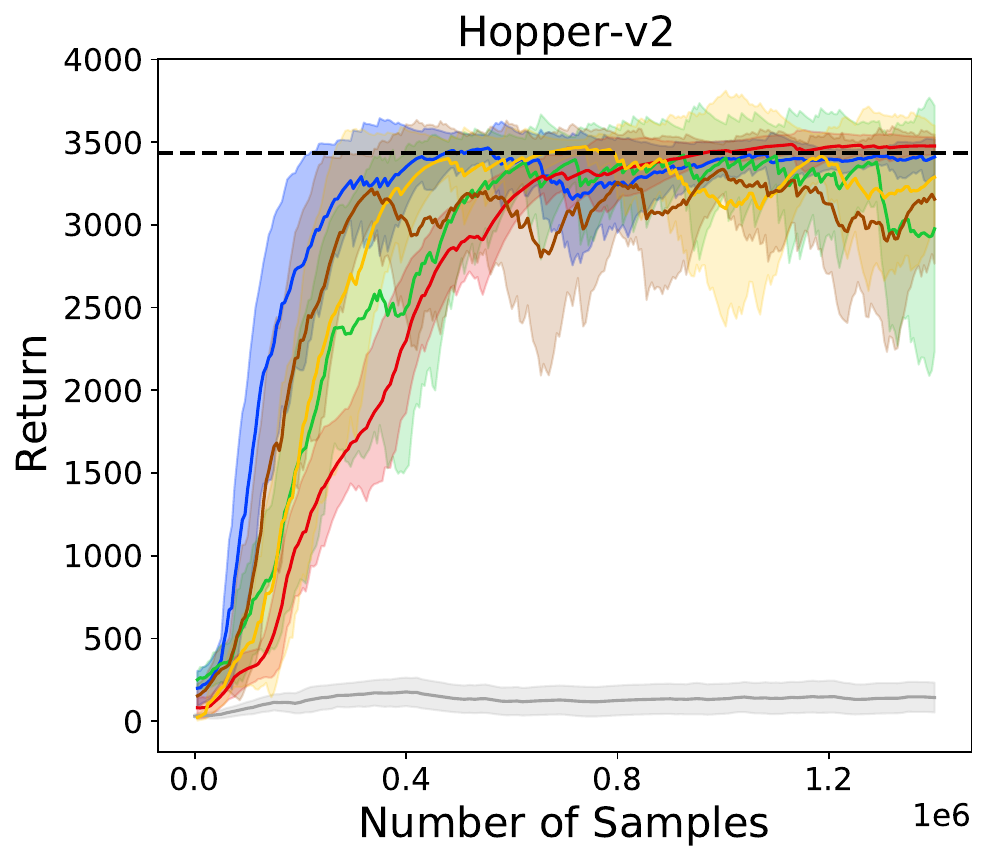}
\end{minipage}
\begin{minipage}{0.32\linewidth}
\vspace{1pt}
\includegraphics[width=\textwidth]{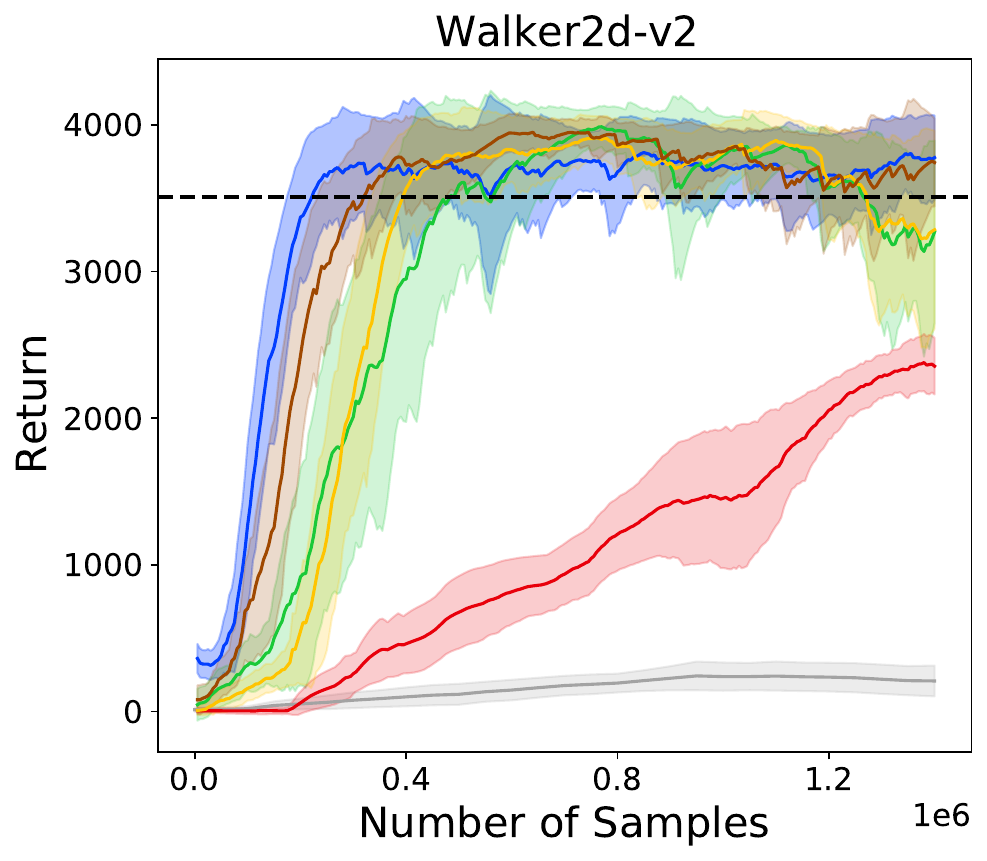}
\end{minipage}
\begin{minipage}{0.32\linewidth}
\vspace{1pt}
\includegraphics[width=\textwidth]{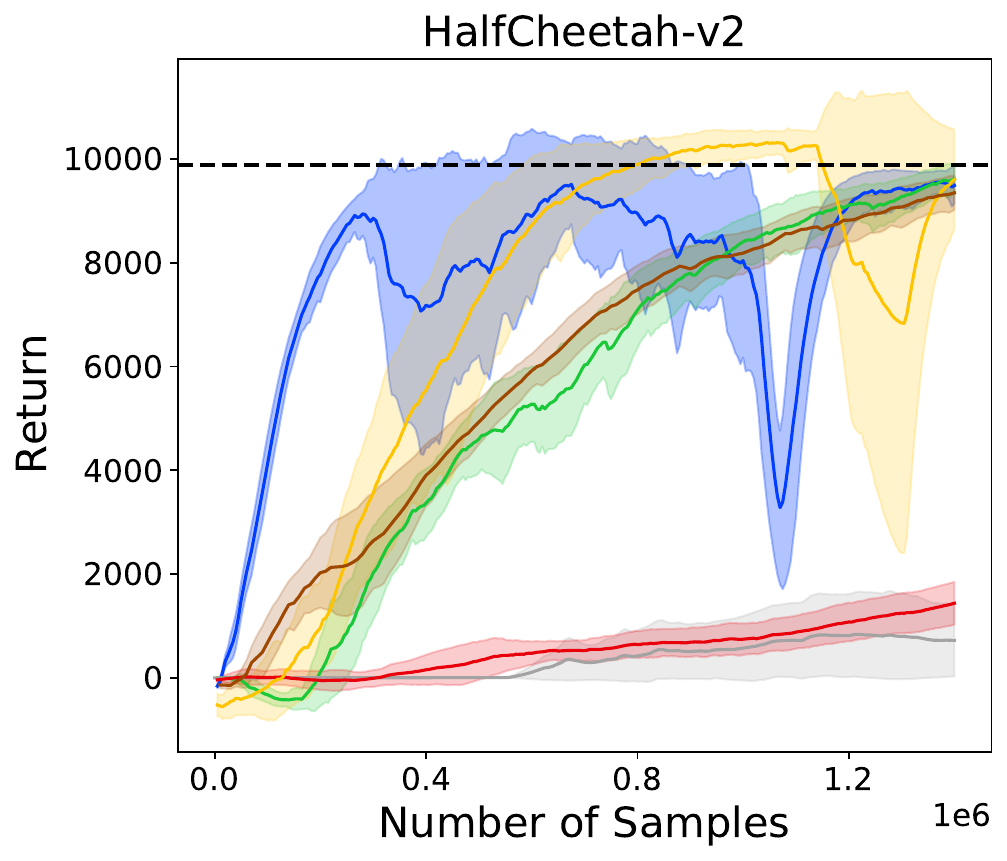}
\end{minipage}
\begin{minipage}{0.96\linewidth}
\vspace{1pt}
\centering
\includegraphics[width=\textwidth]{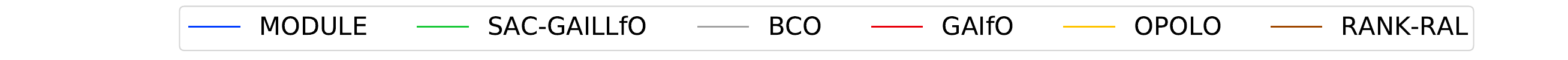}
\end{minipage}
\caption{Comparison of MODULE against advanced LfO methods in three MuJoCo environments. }
\label{experiment_methods_comparison_pic}
\end{figure*}

For the MODULE algorithm, we utilize the multi-layer perceptron (MLP) with two hidden layers to build all networks. Additionally, two independent quantile value networks are employed to mitigate the issue of overestimation. Specifically, in this subsection, IQN is adopted to generate quantile fractions. In our comparisons, we assess MODULE with three LfO baseline methods. To examine whether the distributional RL technique in MODULE mitigates the instability associated with the traditional off-policy scheme, we first implement SAC as the policy training strategy for LfO with adversarial learning, denoted as ``SAC-GAILLfO'' in Fig.\,\ref{experiment_methods_comparison_pic}. Additionally, the comparative methodologies include the following algorithms: BCO \citep{torabi2018behavioral}, GAIfO \citep{torabi2018generative}, OPOLO \citep{zhu2020off} and RANK-RAL \citep{sikchi2023ranking}. 

Fig.\,\ref{experiment_methods_comparison_pic} exhibits the performance of MODULE and all comparative LfO baseline algorithms. We use a dashed line to represent the average return of the expert observation data. The solid line illustrates the mean return of each method, while the shaded area signifies the corresponding standard deviation. Notably, the MODULE algorithm exhibits greater efficiency and stability in learning from expert observations compared to other methods. We attribute this superior performance to two primary factors: (1) the extensive exploration facilitated by the maximum entropy objective, and (2) the stability provided by distributional RL. 

\subsection{Impact of quantile fraction generation methods and risk measures\label{section_experiment_ablation}}
Next, we evaluate the impact of various quantile fraction generation methods and risk measure functions on the performance of MODULE.

\begin{figure*}[htbp]
\centering
\begin{minipage}{0.32\linewidth}
\vspace{1pt}
\includegraphics[width=\textwidth]{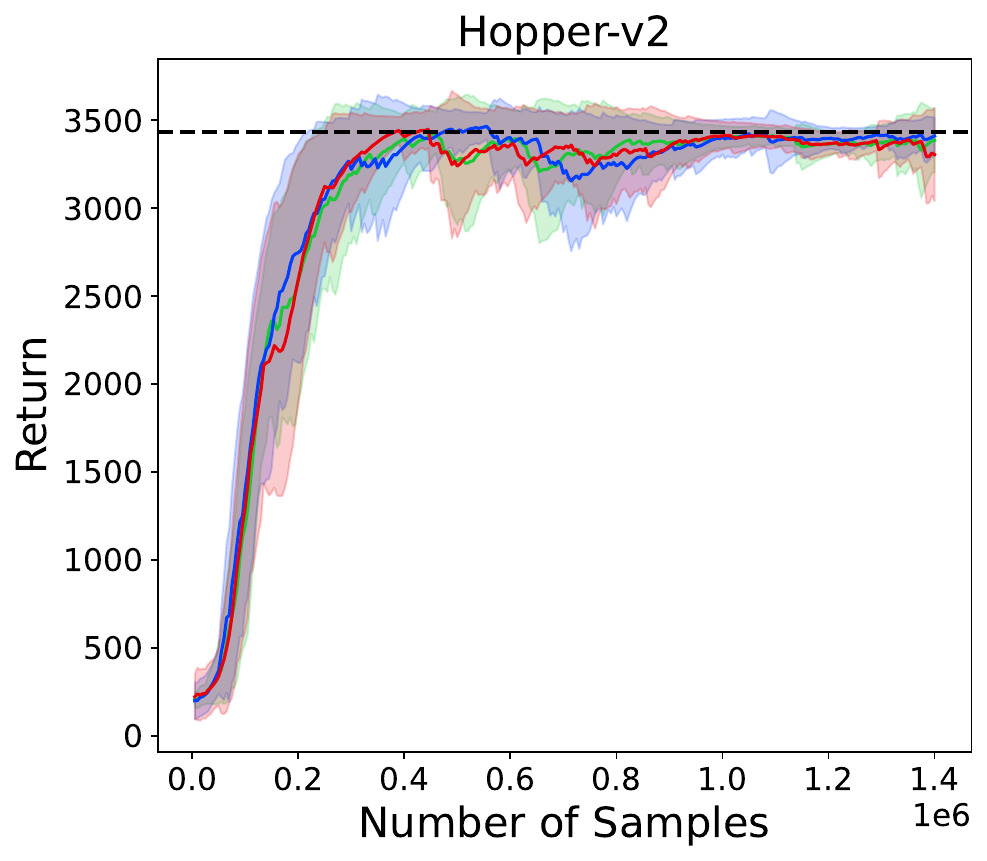}
\end{minipage}
\begin{minipage}{0.32\linewidth}
\vspace{1pt}
\includegraphics[width=\textwidth]{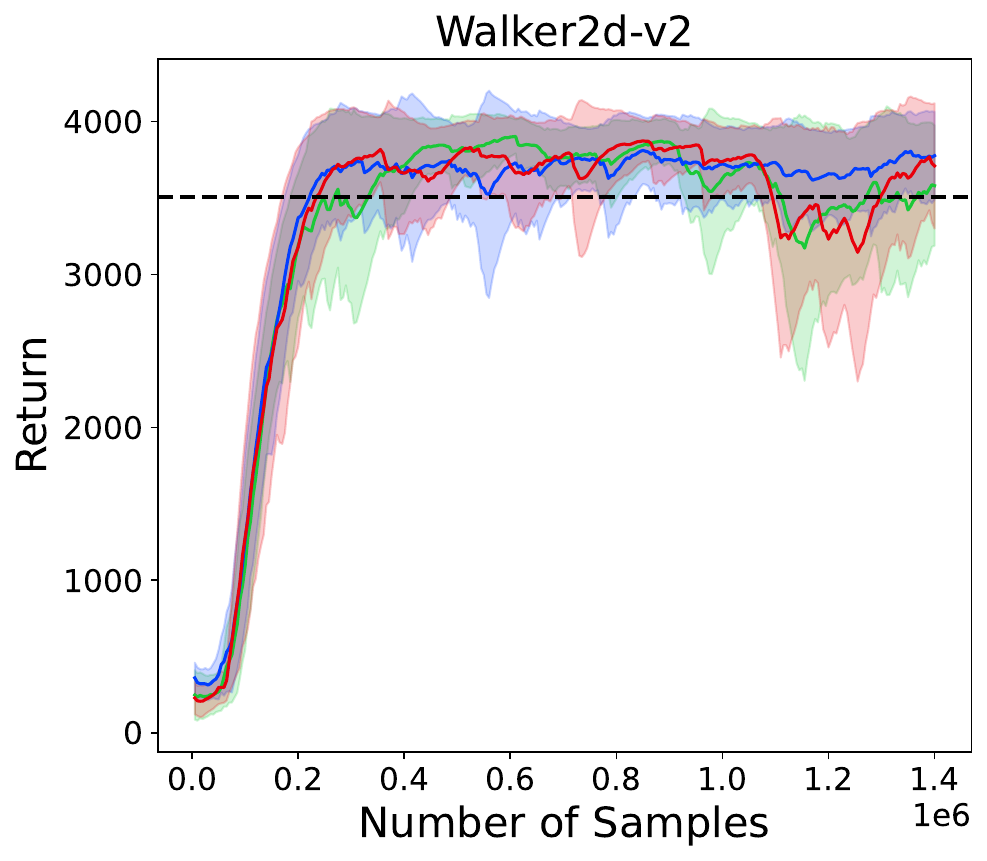}
\end{minipage}
\begin{minipage}{0.32\linewidth}
\vspace{1pt}
\includegraphics[width=\textwidth]{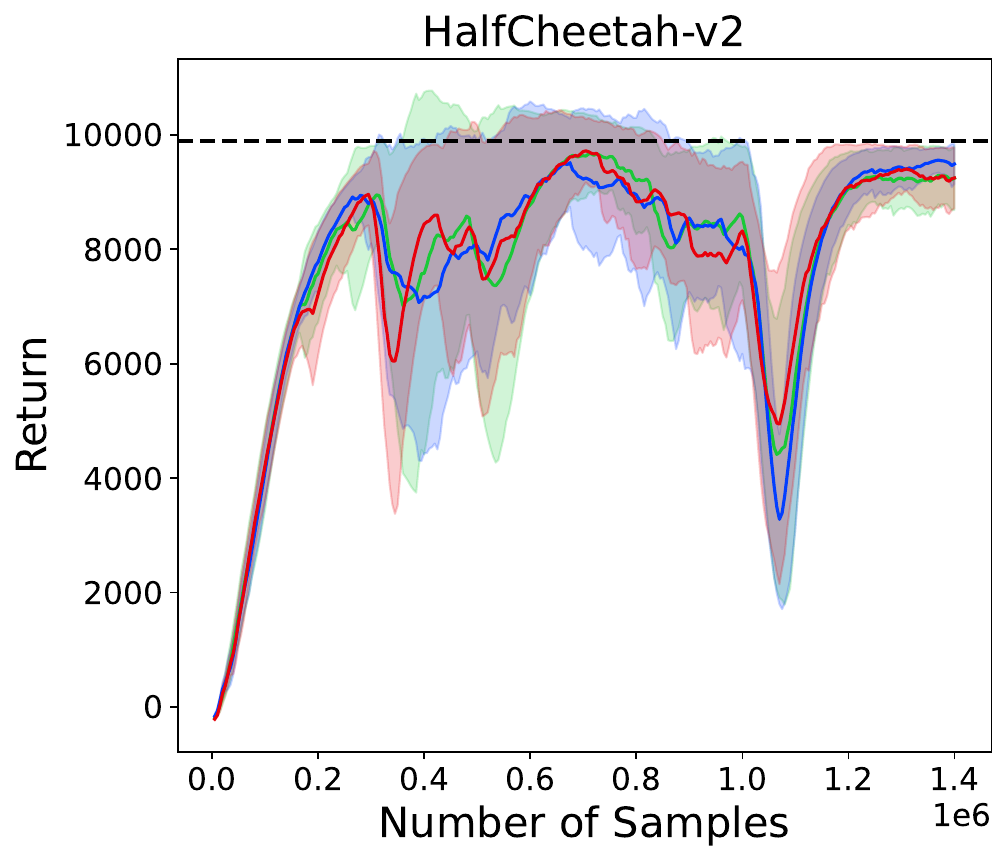}
\end{minipage}
\begin{minipage}{0.96\linewidth}
\vspace{1pt}
\includegraphics[width=\textwidth]{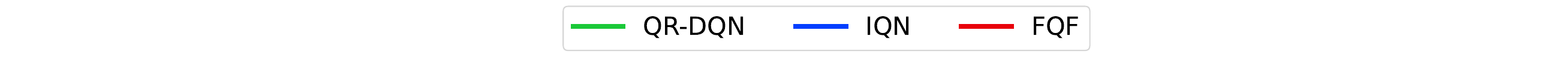}
\end{minipage}
\caption{Performance of three quantile fraction generation methods under the MODULE algorithm in three MuJoCo environments. }
\label{quantile_fraction_generation}
\end{figure*}

Within the MODULE algorithm, we employ three quantile fraction generation methods: QR-DQN, IQN and FQF. Specifically, the quantile proposal network in FQF is a two-layer fully connected network with 128 units and a learning rate of 1e-5 \citep{ma2020dsac}. Regarding the relationship between the quantile value function and the soft Q-function, we employ the risk-neutral measure function. As shown in Fig.\,\ref{quantile_fraction_generation}, all three methods exhibit relative stability in the context of MODULE.

\begin{figure*}[htbp]
\centering
\begin{minipage}{0.32\linewidth}
\vspace{1pt}
\includegraphics[width=\textwidth]{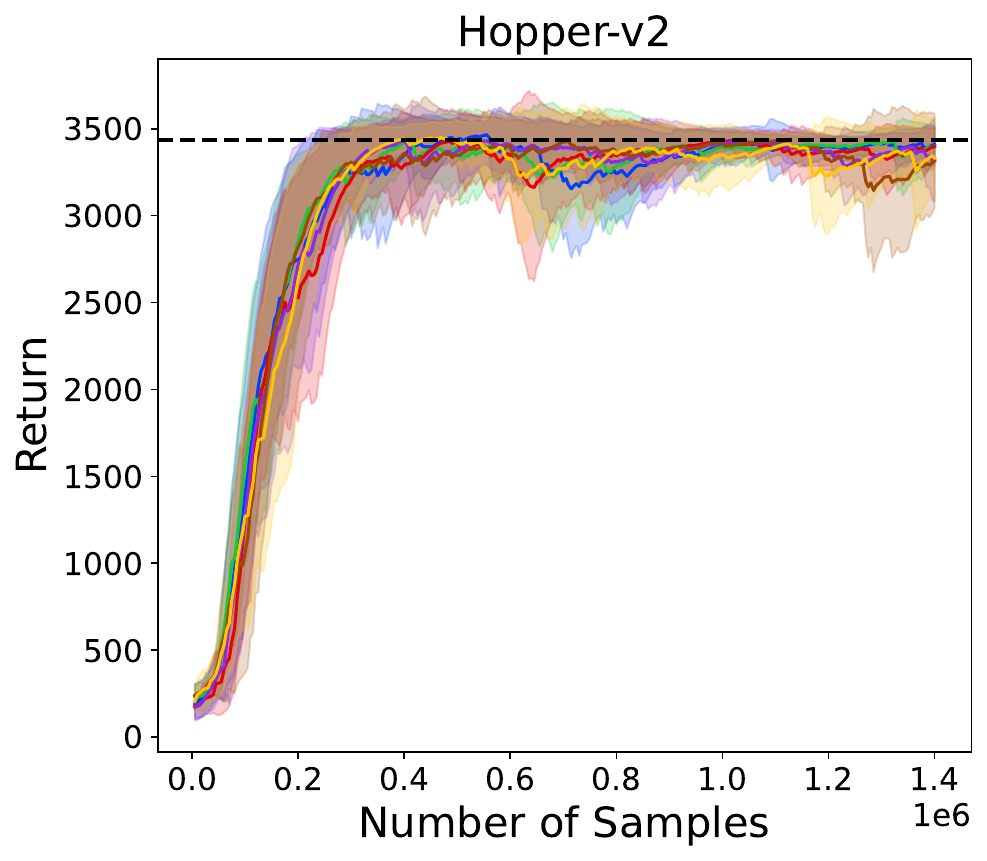}
\end{minipage}
\begin{minipage}{0.32\linewidth}
\vspace{1pt}
\includegraphics[width=\textwidth]{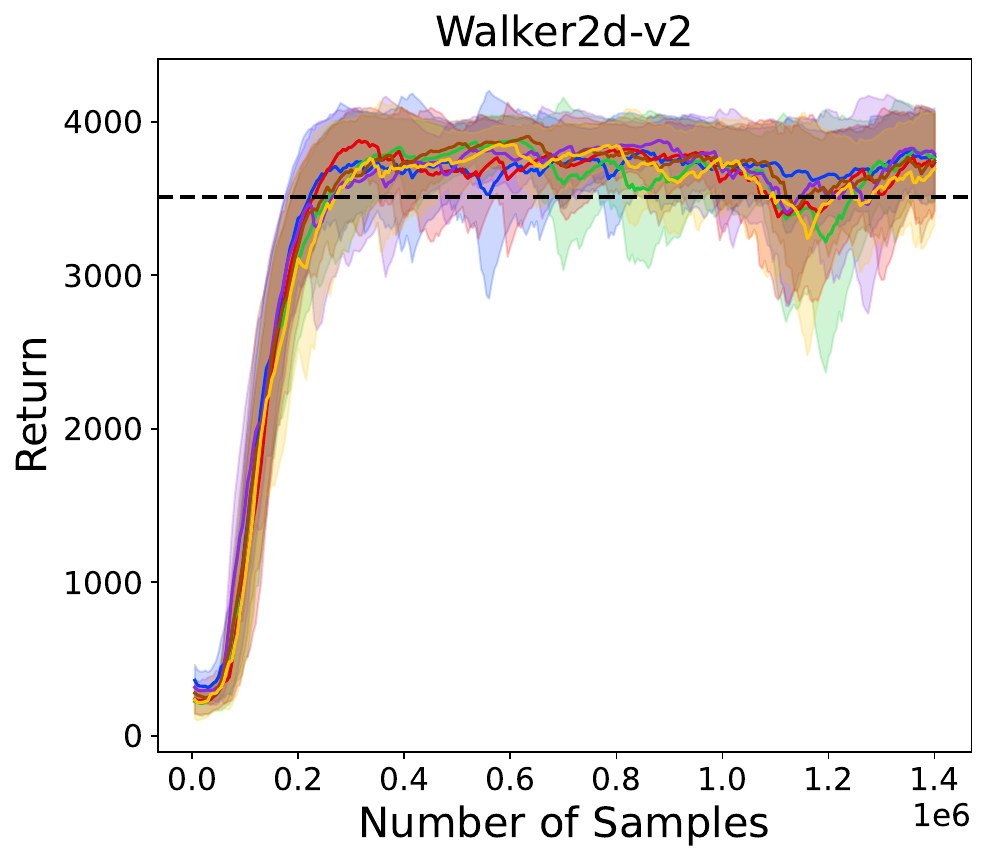}
\end{minipage}
\begin{minipage}{0.32\linewidth}
\vspace{1pt}
\includegraphics[width=\textwidth]{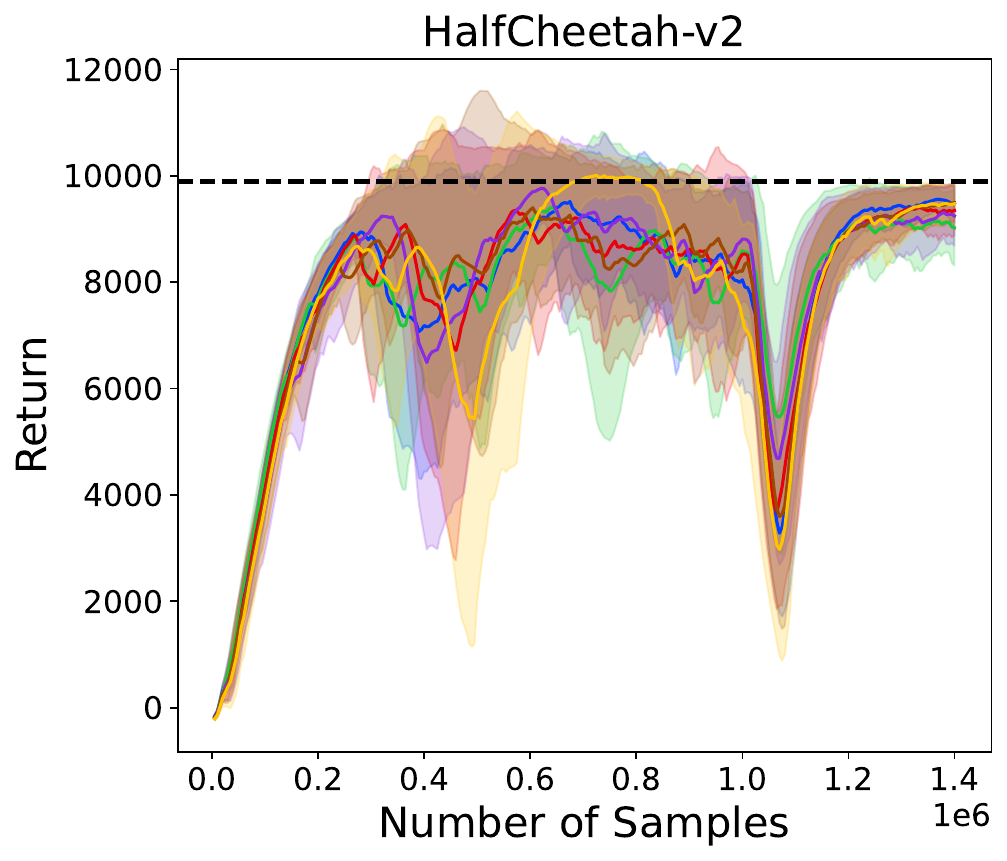}
\end{minipage}
\begin{minipage}{0.96\linewidth}
\vspace{1pt}
\includegraphics[width=\textwidth]{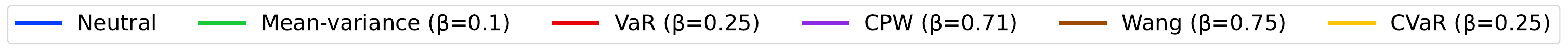}
\end{minipage}
\caption{Performance of five risk-averse measure functions, together with the risk-neutral measure function under the MODULE algorithm in three MuJoCo environments. }
\label{risk_averse}
\end{figure*}

\begin{figure*}[htbp]
\centering
\begin{minipage}{0.32\linewidth}
\vspace{1pt}
\includegraphics[width=\textwidth]{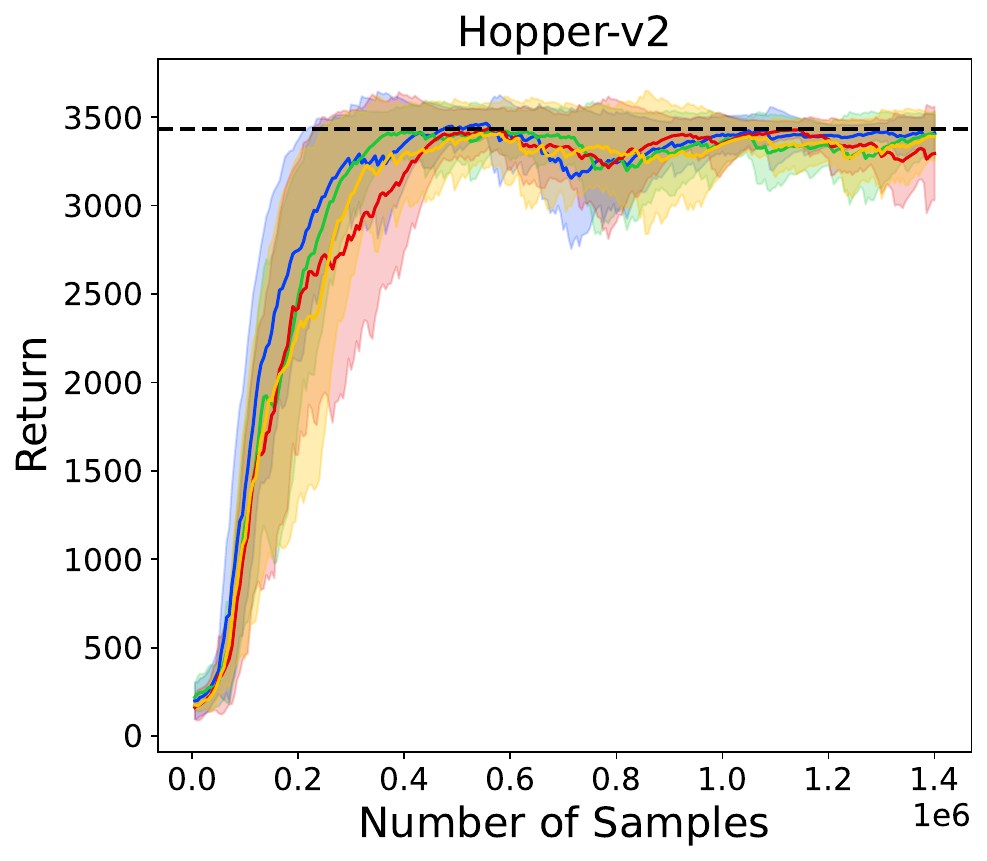}
\end{minipage}
\begin{minipage}{0.32\linewidth}
\vspace{1pt}
\includegraphics[width=\textwidth]{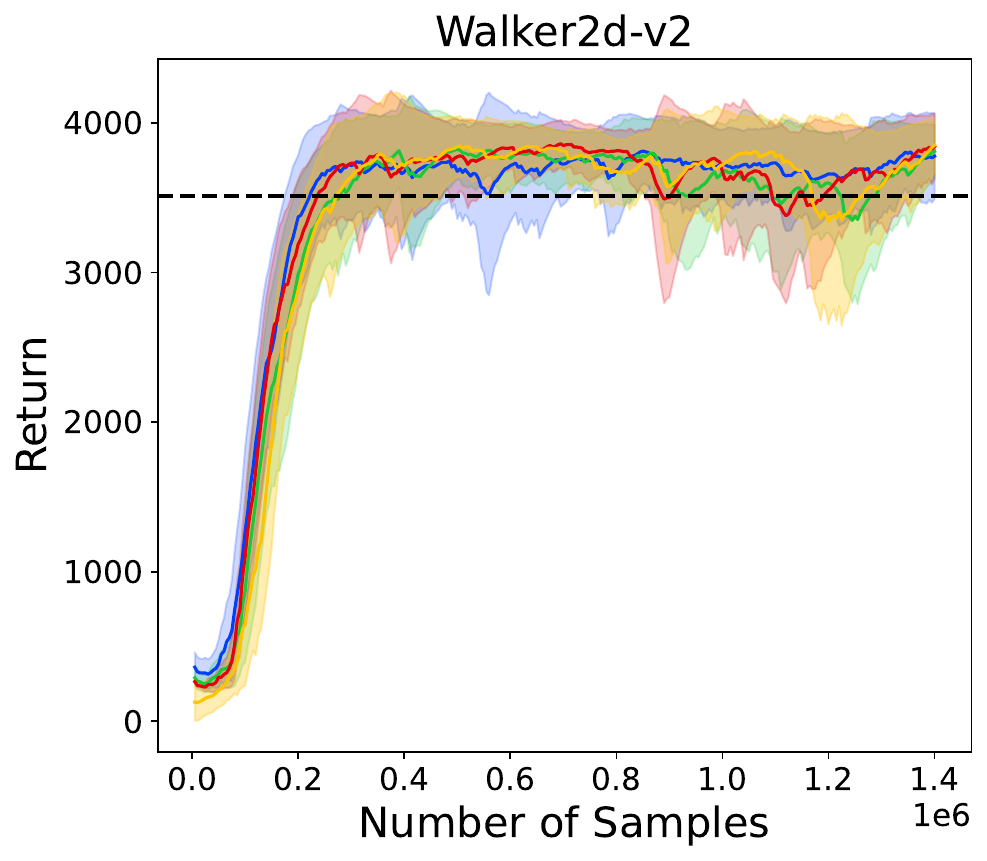}
\end{minipage}
\begin{minipage}{0.32\linewidth}
\vspace{1pt}
\includegraphics[width=\textwidth]{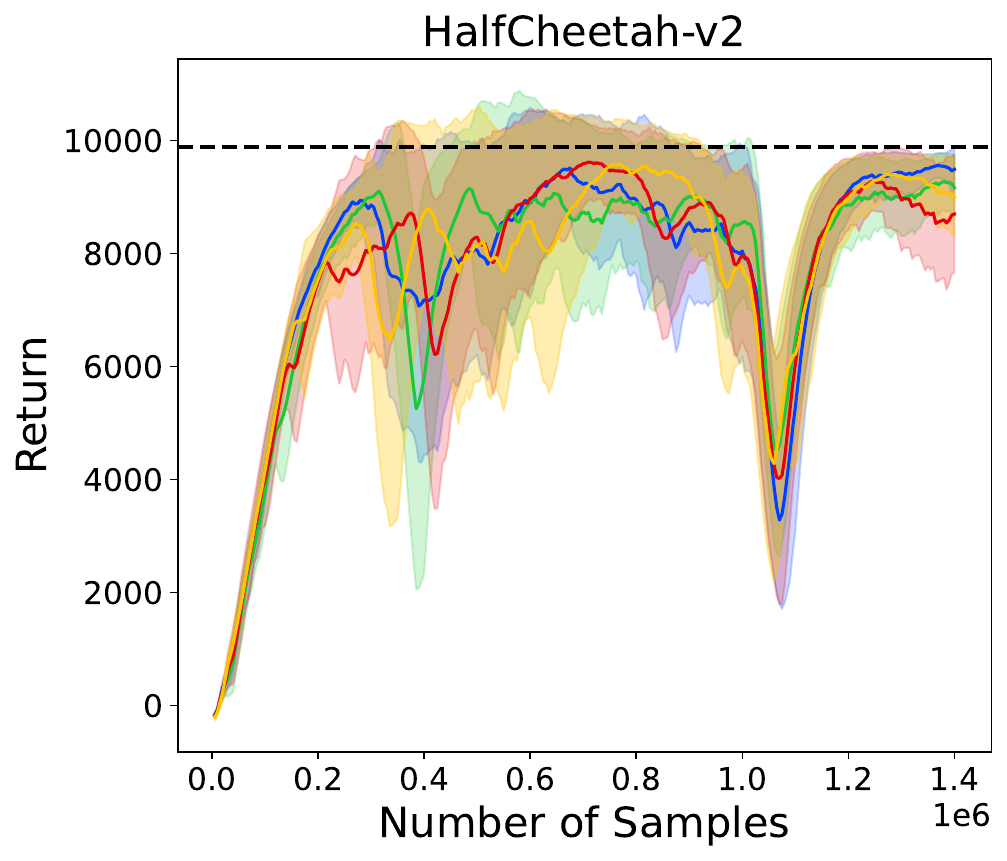}
\end{minipage}
\begin{minipage}{0.96\linewidth}
\vspace{1pt}
\includegraphics[width=\textwidth]{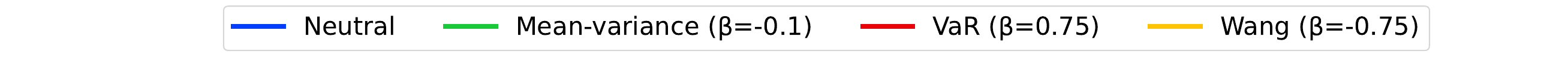}
\end{minipage}
\caption{Performance of three risk-seeking measure functions, together with the risk-neutral measure function under the MODULE algorithm in three MuJoCo environments. }
\label{risk_seeking}
\end{figure*}

For the risk measure functions, we consider the approaches mentioned in Section \ref{section_MODULE} under the MODULE algorithm. Particularly, we adopt identical distributional RL configurations for each risk measure function as outlined in \citep{dabney2018implicit,ma2020dsac}. 
\begin{itemize}
\item Five risk-averse learned policies are accessed. For mean-variance, $\beta$ is set as 0.1. For VaR, $\beta$ is set as 0.25. For CPW, $\beta$ is set as 0.71. For Wang, $\beta$ is set as 0.75. For CVaR, $\beta$ is set as 0.25. 
\item Three risk-seeking learned policies are accessed. For mean-variance, $\beta$ is set as -0.1. For VaR, $\beta$ is set as 0.75. For Wang, $\beta$ is set as -0.75. 
\end{itemize}
As exhibited in Figs.\,\ref{risk_averse} and \ref{risk_seeking}, MODULE demonstrates relatively stable performance, indicating that the risk measure functions are insensitive within the MODULE algorithm. 

\section{Conclusion and discussion\label{section_conclusion}}
This paper investigates the generalization and computational properties of the LfO problem. We begin by examining the generalization capabilities of the reward function and policy in LfO. Subsequently, we introduce the MODULE algorithm, which leverages the strengths of SAC and distributional RL. Particularly, the SAC component of MODULE enhances sample efficiency and training robustness, while the distributional RL component improves training stability. Extensive experiments in MuJoCo environments demonstrate that MODULE excels at imitating observational data without explicit action signals. 

In addition to adversarial training, OT is also employed for reward calculation \citep{arjovsky2017wasserstein,papagiannis2022imitation,luo2023optimal,liu2024imitation}. By integrating distributional RL with OT methods, modeling the uncertainty in the reward distribution has the potential to enhance the stability of the training process, especially in environments characterized by sparse or noisy reward signals. We regard this area to be explored in the future. 

\section*{Acknowledgments}
This work was supported in part by the National Natural Science Foundation of China under Grant 12301351. 

\newpage
\section*{Appendix}
\appendix
\section{Proofs for LfO generalization properties}
\setcounter{equation}{0}
\renewcommand{\theequation}{A.\arabic{equation}}
\subsection{Proof of Theorem \ref{theorem_lfo_generalization_reward}\label{proof_theorem_lfo_generalization_reward}}
\begin{proof}
Recall the definition of the LfO reward distance (Definition \ref{definition_LfO_reward_distance}), we have that
\begin{align}
&~~~~d_{\mathcal{R}}^{\rm LfO}(\mu_{\pi_{\rm E}},\mu_{\pi_{\rm I}}) - d_{\mathcal{R}}^{\rm LfO}(\hat{\mu}_{\pi_{\rm E}},\hat{\mu}_{\pi_{\rm I}}) \notag\\
&=\sup_{r\in \mathcal{R}}\left\{ \mathbb{E}_{(s,s')\sim \mu_{\pi_{\rm E}}}{[r(s,s')]}-\mathbb{E}_{(s,s')\sim \mu_{\pi_{\rm I}}}{[r(s,s')]}\right\} - \sup_{r\in \mathcal{R}}\left\{ \mathbb{E}_{(s,s')\sim \hat{\mu}_{\pi_{\rm E}}}{[r(s,s')]}-\mathbb{E}_{(s,s')\sim \hat{\mu}_{\pi_{\rm I}}}{[r(s,s')]}\right\} \notag\\
&\leq \sup_{r\in \mathcal{R}}\Big\{ \big( \mathbb{E}_{(s,s')\sim \mu_{\pi_{\rm E}}}{[r(s,s')]}-\mathbb{E}_{(s,s')\sim \mu_{\pi_{\rm I}}}{[r(s,s')]} \big) -\big( \mathbb{E}_{(s,s')\sim \hat{\mu}_{\pi_{\rm E}}}{[r(s,s')]}-\mathbb{E}_{(s,s')\sim \hat{\mu}_{\pi_{\rm I}}}{[r(s,s')]} \big) \Big\} \notag\\
&\leq \sup_{r\in \mathcal{R}}\left\{ \mathbb{E}_{(s,s')\sim \mu_{\pi_{\rm E}}}{[r(s,s')]}- \mathbb{E}_{(s,s')\sim \hat{\mu}_{\pi_{\rm E}}}{[r(s,s')]} \right\} + \sup_{r\in \mathcal{R}} \left\{ \mathbb{E}_{(s,s')\sim \hat{\mu}_{\pi_{\rm I}}}{[r(s,s')]} - \mathbb{E}_{(s,s')\sim \mu_{\pi_{\rm I}}}{[r(s,s')]} \right\} \notag\\
&\leq \sup_{r\in \mathcal{R}}\left| \mathbb{E}_{(s,s')\sim \mu_{\pi_{\rm E}}}{[r(s,s')]}- \mathbb{E}_{(s,s')\sim \hat{\mu}_{\pi_{\rm E}}}{[r(s,s')]} \right| + \sup_{r\in \mathcal{R}} \left| \mathbb{E}_{(s,s')\sim \mu_{\pi_{\rm I}}}{[r(s,s')]} - \mathbb{E}_{(s,s')\sim \hat{\mu}_{\pi_{\rm I}}}{[r(s,s')]} \right|. 
\label{th1_relationship_distance_from_empirical}
\end{align}

We first focus on the boundedness of
$$
\sup_{r\in \mathcal{R}}\left| \mathbb{E}_{(s,s')\sim \mu_{\pi_{\rm E}}}{[r(s,s')]}- \mathbb{E}_{(s,s')\sim \hat{\mu}_{\pi_{\rm E}}}{[r(s,s')]} \right|. 
$$
Employing McDiarmid's inequality \citep{mohri2018foundations}, then with probability at least $1-\delta/4$, we obtain that
\begin{align}
&~~~~\sup_{r\in \mathcal{R}}\left| \mathbb{E}_{(s,s')\sim \mu_{\pi_{\rm E}}}{[r(s,s')]}- \mathbb{E}_{(s,s')\sim \hat{\mu}_{\pi_{\rm E}}}{[r(s,s')]} \right| \notag\\
&\leq \mathbb{E} \left[ \sup_{r\in \mathcal{R}}\left| \mathbb{E}_{(s,s')\sim \mu_{\pi_{\rm E}}}{[r(s,s')]}- \mathbb{E}_{(s,s')\sim \hat{\mu}_{\pi_{\rm E}}}{[r(s,s')]} \right| \right] + 2B_{r}\sqrt{\frac{\log(4/\delta)}{2n}}, 
\label{th1_pi_E_add_expectation}
\end{align}
where the external expectation is over the random sampling of $\hat{\mu}_{\pi_{\rm E}}$ with $n$ state transition pairs. Based on the Rademacher complexity theory \citep{mohri2018foundations}, the first term in Eq. \eqref{th1_pi_E_add_expectation} can be bounded by
\begin{align}
&~~~~\mathbb{E} \left[ \sup_{r\in \mathcal{R}}\left| \mathbb{E}_{(s,s')\sim \mu_{\pi_{\rm E}}}{[r(s,s')]}- \mathbb{E}_{(s,s')\sim \hat{\mu}_{\pi_{\rm E}}}{[r(s,s')]} \right| \right] \notag\\
&\leq 2\mathbb{E}_{\boldsymbol{\sigma},\mu_{\pi_{\rm E}}}\left[\sup_{r\in \mathcal{R}}\sum_{i=1}^n\frac{1}{n}\sigma_i r(s^{(i)},s'^{(i)})\right] \notag\\
&=2\mathfrak{R}_{\mu_{\pi_{\rm E}}}^{(n)}(\mathcal{R}).
\label{th1_expectation_to_Rademacher_complexity}
\end{align}
Connecting the Rademacher complexity to its empirical version, then with probability at least $1-\delta/4$, we get that
\begin{align}
\mathfrak{R}_{\mu_{\pi_{\rm E}}}^{(n)}(\mathcal{R})\leq \hat{\mathfrak{R}}_{\mu_{\pi_{\rm E}}}^{(n)}(\mathcal{R}) + 2B_{r}\sqrt{\frac{\log(4/\delta)}{2n}}, 
\label{th1_Rademacher_complexity_to_empirical}
\end{align}
where 
\begin{align*}
\hat{\mathfrak{R}}_{\mu_{\pi_{\rm E}}}^{(n)}(\mathcal{R})=\mathbb{E}_{\boldsymbol{\sigma}}\left[\sup_{r\in \mathcal{R}}\sum_{i=1}^n\frac{1}{n}\sigma_i r(s^{(i)},s'^{(i)})\right]. 
\end{align*}
Combining Eq. \eqref{th1_pi_E_add_expectation} with Eqs. \eqref{th1_expectation_to_Rademacher_complexity} and \eqref{th1_Rademacher_complexity_to_empirical}, with probability at least $1-\delta/2$, we obtain that
\begin{align}
\sup_{r\in \mathcal{R}}\left| \mathbb{E}_{(s,s')\sim \mu_{\pi_{\rm E}}}{[r(s,s')]}- \mathbb{E}_{(s,s')\sim \hat{\mu}_{\pi_{\rm E}}}{[r(s,s')]} \right| 
\leq 2\hat{\mathfrak{R}}_{\mu_{\pi_{\rm E}}}^{(n)}(\mathcal{R}) + 6B_{r}\sqrt{\frac{\log(4/\delta)}{2n}}. 
\label{th1_pi_E_bound}
\end{align}

Similarly, for the boundedness of
\begin{align*}
\sup_{r\in \mathcal{R}} \left| \mathbb{E}_{(s,s')\sim \mu_{\pi_{\rm I}}}{[r(s,s')]} - \mathbb{E}_{(s,s')\sim \hat{\mu}_{\pi_{\rm I}}}{[r(s,s')]} \right|, 
\end{align*}
we have that 
\begin{align}
\sup_{r\in \mathcal{R}}\left| \mathbb{E}_{(s,s')\sim \mu_{\pi_{\rm I}}}{[r(s,s')]}- \mathbb{E}_{(s,s')\sim \hat{\mu}_{\pi_{\rm I}}}{[r(s,s')]} \right| 
\leq 2\hat{\mathfrak{R}}_{\mu_{\pi_{\rm I}}}^{(n)}(\mathcal{R}) + 6B_{r}\sqrt{\frac{\log(4/\delta)}{2n}}. 
\label{th1_pi_I_bound}
\end{align}

Combining Eq. \eqref{th1_relationship_distance_from_empirical} with Eqs. \eqref{th1_pi_E_bound} and \eqref{th1_pi_I_bound} completes the proof. 
\end{proof}

\subsection{Proof of Theorem \ref{theorem_lfo_generalization_policy}\label{proof_theorem_lfo_generalization_policy}}
\begin{proof}
Recall the definition of the state transition distribution error (Definition \ref{definition_state_transition_distribution_error}), we have that
\begin{align}
&\hskip-0.7cm ~~~~\bm{e}(C_{r_{\rm I}}\mu_{\pi_{\rm E}},C_{r_{\rm I}}\mu_{\Pi})-\bm{e}(\hat{c}_{r_{\rm I}}^{(m)}\mu_{\pi_{\rm E}},\hat{c}_{r_{\rm I}}^{(m)}\mu_{\Pi}) \notag\\
&\hskip-0.7cm =C_{r_{\rm I}}\inf_{\pi \in \Pi}\left\{ \mathbb{E}_{(s,s')\sim \mathbb{R}_{\rm I}}\big[ \mu_{\pi_{\rm E}}(s,s')-\mu_{\pi}(s,s') \big]\right\} -\hat{c}_{r_{\rm I}}^{(m)}\inf_{\pi \in \Pi}\left\{ \mathbb{E}_{(s,s')\sim \hat{\mathbb{R}}_{\rm I}}\big[ \mu_{\pi_{\rm E}}(s,s')-\mu_{\pi}(s,s') \big]\right\} \notag\\
&\hskip-0.7cm \overset{(\romannumeral1)}{\geq} C_{r_{\rm I}}\inf_{\pi \in \Pi}\left\{ \mathbb{E}_{(s,s')\sim \mathbb{R}_{\rm I}}\big[ \mu_{\pi_{\rm E}}(s,s')-\mu_{\pi}(s,s') \big]\right\} -C_{r_{\rm I}}\inf_{\pi \in \Pi}\left\{ \mathbb{E}_{(s,s')\sim \hat{\mathbb{R}}_{\rm I}}\big[ \mu_{\pi_{\rm E}}(s,s')-\mu_{\pi}(s,s') \big]\right\} \notag\\
&\hskip-0.7cm \geq C_{r_{\rm I}}\inf_{\pi \in \Pi}\Big\{ \mathbb{E}_{(s,s')\sim \mathbb{R}_{\rm I}}\big[ \mu_{\pi_{\rm E}}(s,s')-\mu_{\pi}(s,s') \big] -\mathbb{E}_{(s,s')\sim \hat{\mathbb{R}}_{\rm I}}\big[ \mu_{\pi_{\rm E}}(s,s')-\mu_{\pi}(s,s') \big]\Big\} \notag\\
&\hskip-0.7cm \geq C_{r_{\rm I}}\inf_{\pi \in \Pi}\Big\{ \mathbb{E}_{(s,s')\sim \mathbb{R}_{\rm I}}\big[ \mu_{\pi_{\rm E}}(s,s') \big] -\mathbb{E}_{(s,s')\sim \hat{\mathbb{R}}_{\rm I}}\big[ \mu_{\pi_{\rm E}}(s,s') \big]\Big\} +C_{r_{\rm I}}\inf_{\pi \in \Pi}\Big\{ \mathbb{E}_{(s,s')\sim \hat{\mathbb{R}}_{\rm I}}\big[ \mu_{\pi}(s,s') \big] -\mathbb{E}_{(s,s')\sim \mathbb{R}_{\rm I}}\big[ \mu_{\pi}(s,s') \big]\Big\} \notag\\
&\hskip-0.7cm =\Big( \mathbb{E}_{(s,s')\sim \mathbb{R}_{\rm I}}\big[ C_{r_{\rm I}}\mu_{\pi_{\rm E}}(s,s') \big]-\mathbb{E}_{(s,s')\sim \hat{\mathbb{R}}_{\rm I}}\big[ C_{r_{\rm I}}\mu_{\pi_{\rm E}}(s,s') \big] \Big) -\sup_{\pi \in \Pi}\left\{ \mathbb{E}_{(s,s')\sim \mathbb{R}_{\rm I}}\big[ C_{r_{\rm I}}\mu_{\pi}(s,s') \big]-\mathbb{E}_{(s,s')\sim \hat{\mathbb{R}}_{\rm I}}\big[ C_{r_{\rm I}}\mu_{\pi}(s,s') \big] \right\}, 
\label{th2_relationship_error_from_empirical}
\end{align}
where (\romannumeral1) comes from $\hat{c}_{r_{\rm I}}^{(m)}\leq C_{r_{\rm I}}$ and the non-negative property of $\bm{e}(\hat{c}_{r_{\rm I}}^{(m)}\mu_{\pi_{\rm E}},\hat{c}_{r_{\rm I}}^{(m)}\mu_{\Pi})$. 

We first focus on the boundedness of 
$$
\mathbb{E}_{(s,s')\sim \mathbb{R}_{\rm I}}\big[ C_{r_{\rm I}}\mu_{\pi_{\rm E}}(s,s') \big]-\mathbb{E}_{(s,s')\sim \hat{\mathbb{R}}_{\rm I}}\big[ C_{r_{\rm I}}\mu_{\pi_{\rm E}}(s,s') \big]. 
$$
Employing McDiarmid's inequality, then with probability at least $1-\delta/3$, we have that
\begin{align*}
&~~~~\mathbb{E}_{(s,s')\sim \hat{\mathbb{R}}_{\rm I}}\big[ C_{r_{\rm I}}\mu_{\pi_{\rm E}}(s,s') \big] \\
&\leq \mathbb{E}\left[ \mathbb{E}_{(s,s')\sim \hat{\mathbb{R}}_{\rm I}}\big[ C_{r_{\rm I}}\mu_{\pi_{\rm E}}(s,s') \big] \right]+2B_{\Pi}\sqrt{\frac{\log(3/\delta)}{2m}} \\
&=\mathbb{E}_{(s,s')\sim \mathbb{R}_{\rm I}}\big[ C_{r_{\rm I}}\mu_{\pi_{\rm E}}(s,s') \big]+2B_{\Pi}\sqrt{\frac{\log(3/\delta)}{2m}}, 
\end{align*}
i.e.,
\begin{align}
\mathbb{E}_{(s,s')\sim \mathbb{R}_{\rm I}}\big[ C_{r_{\rm I}}\mu_{\pi_{\rm E}}(s,s') \big] - \mathbb{E}_{(s,s')\sim \hat{\mathbb{R}}_{\rm I}}\big[ C_{r_{\rm I}}\mu_{\pi_{\rm E}}(s,s') \big]
\geq -2B_{\Pi}\sqrt{\frac{\log(3/\delta)}{2m}}. 
\label{th2_pi_E_bound}
\end{align}

We then investigate the boundedness of
$$
\sup_{\pi \in \Pi}\left\{ \mathbb{E}_{(s,s')\sim \mathbb{R}_{\rm I}}\big[ C_{r_{\rm I}}\mu_{\pi}(s,s') \big]-\mathbb{E}_{(s,s')\sim \hat{\mathbb{R}}_{\rm I}}\big[ C_{r_{\rm I}}\mu_{\pi}(s,s') \big] \right\}. 
$$
By McDiarmid's inequality, then with probability at least $1-\delta/3$, we obtain that
\begin{align}
&\hskip-0.9cm ~~~~\sup_{\pi \in \Pi}\left\{ \mathbb{E}_{(s,s')\sim \mathbb{R}_{\rm I}}\big[ C_{r_{\rm I}}\mu_{\pi}(s,s') \big]-\mathbb{E}_{(s,s')\sim \hat{\mathbb{R}}_{\rm I}}\big[ C_{r_{\rm I}}\mu_{\pi}(s,s') \big] \right\} \notag\\
&\hskip-0.9cm \leq \mathbb{E} \left[ \sup_{\pi \in \Pi}\left\{ \mathbb{E}_{(s,s')\sim \mathbb{R}_{\rm I}}\big[ C_{r_{\rm I}}\mu_{\pi}(s,s') \big]-\mathbb{E}_{(s,s')\sim \hat{\mathbb{R}}_{\rm I}}\big[ C_{r_{\rm I}}\mu_{\pi}(s,s') \big] \right\} \right] + 2B_{\Pi}\sqrt{\frac{\log(3/\delta)}{2m}}, 
\label{th2_pi_add_expectation}
\end{align}
where the external expectation is over the random sampling of the distribution $\hat{\mathbb{R}}_{\rm I}$ with $m$ state-transition pairs. Based on the Rademacher complexity theory, the first term in Eq. \eqref{th2_pi_add_expectation} can be bounded by
\begin{align}
&\hskip-0.9cm ~~~~\mathbb{E} \left[ \sup_{\pi \in \Pi}\left\{ \mathbb{E}_{(s,s')\sim \mathbb{R}_{\rm I}}\big[ C_{r_{\rm I}}\mu_{\pi}(s,s') \big]-\mathbb{E}_{(s,s')\sim \hat{\mathbb{R}}_{\rm I}}\big[ C_{r_{\rm I}}\mu_{\pi}(s,s') \big] \right\} \right] \notag\\
&\hskip-0.9cm \leq 2\mathbb{E}_{\boldsymbol{\sigma},\mathbb{R}_{\rm I}}\left[\sup_{\pi \in \Pi}\sum_{j=1}^m\frac{1}{m}\sigma_j C_{r_{\rm I}}\mu_{\pi}(s^{(j)},s'^{(j)})\right] \notag\\
&\hskip-0.9cm =2\mathfrak{R}_{\mathbb{R}_{\rm I}}^{(m)}(C_{r_{\rm I}}\mu_{\Pi}).
\label{th2_expectation_to_Rademacher_complexity}
\end{align}
Connecting the Rademacher complexity to its empirical version, then with probability at least $1-\delta/3$, we get that
\begin{align}
\mathfrak{R}_{\mathbb{R}_{\rm I}}^{(m)}(C_{r_{\rm I}}\mu_{\Pi})\leq \hat{\mathfrak{R}}_{\mathbb{R}_{\rm I}}^{(m)}(C_{r_{\rm I}}\mu_{\Pi}) + 2B_{\Pi}\sqrt{\frac{\log(3/\delta)}{2m}}, 
\label{th2_Rademacher_complexity_to_empirical}
\end{align}
where 
\begin{align*}
\hat{\mathfrak{R}}_{\mathbb{R}_{\rm I}}^{(m)}(C_{r_{\rm I}}\mu_{\Pi})=\mathbb{E}_{\boldsymbol{\sigma}}\left[\sup_{\pi \in \Pi}\sum_{j=1}^m\frac{1}{m}\sigma_j C_{r_{\rm I}}\mu_{\pi}(s^{(j)},s'^{(j)})\right]. 
\end{align*}
Combining Eq. \eqref{th2_pi_add_expectation} with Eqs. \eqref{th2_expectation_to_Rademacher_complexity} and \eqref{th2_Rademacher_complexity_to_empirical}, with probability at least $1-2\delta/3$, we obtain that
\begin{align}
&\hskip-0.9cm ~~~~\sup_{\pi \in \Pi}\left\{ \mathbb{E}_{(s,s')\sim \mathbb{R}_{\rm I}}\big[ C_{r_{\rm I}}\mu_{\pi}(s,s') \big]-\mathbb{E}_{(s,s')\sim \hat{\mathbb{R}}_{\rm I}}\big[ C_{r_{\rm I}}\mu_{\pi}(s,s') \big] \right\} 
\leq 2\hat{\mathfrak{R}}_{\mathbb{R}_{\rm I}}^{(m)}(C_{r_{\rm I}}\mu_{\Pi}) + 6B_{\Pi}\sqrt{\frac{\log(3/\delta)}{2m}}. 
\label{th2_pi_bound}
\end{align}

Combining Eq. \eqref{th2_relationship_error_from_empirical} with Eqs. \eqref{th2_pi_E_bound} and \eqref{th2_pi_bound} completes the proof. 
\end{proof}

\bibliographystyle{apalike}
\bibliography{ref}

\end{document}